\documentclass[preprint,review,3p,times,a4paper]{elsarticle}
\usepackage{amssymb}
\usepackage[linesnumbered,ruled,vlined, ruled,vlined]{algorithm2e}
\usepackage{amsmath}
\usepackage{lineno}
\usepackage[miktex]{gnuplottex}
\usepackage{pgfplots}
\pgfplotsset{compat=1.18}
\usepackage{subcaption}
\usepackage{tikz}
\usepackage{xcolor}
\usepgfplotslibrary{fillbetween}
\usepackage{float}
\usepackage{soul}

\def\ie{i.e., }

\journal{Fluids}
\begin{document}
\begin{frontmatter}
\title{A Physics-Guided Bi-Fidelity Fourier-Featured Operator Learning Framework for Predicting Time Evolution of Drag and Lift Coefficients}
\author[label2]{Amirhossein Mollaali}
\author[label2]{Izzet Sahin}
\author[label3]{Iqrar Raza}
\author[label3]{Christian Moya}
\author[label2]{Guillermo Paniagua}
\author[label2,label3]{Guang Lin\corref{core}}
	\ead{guanglin@purdue.edu}
    \cortext[core]{corresponding author}
 \affiliation[label2]{organization={Purdue University},
             addressline={School of Mechanical Engineering},
             city={West Lafayette},
             postcode={47906},
             state={IN},
             country={USA}}
 \affiliation[label3]{organization={Purdue University},
			 addressline={Department of Mathematics},
		 	 city={West Lafayette},
	 		 postcode={47906},
			 state={IN},
			 country={USA}}
\begin{abstract}
In the pursuit of accurate experimental and computational data while minimizing effort, there is a constant need for high-fidelity results. However, achieving such results often requires significant computational resources. To address this challenge, this paper proposes a deep operator learning-based framework that requires a limited high-fidelity dataset for training. We introduce a novel physics-guided, bi-fidelity, Fourier-featured Deep Operator Network (DeepONet) framework that effectively combines low and high-fidelity datasets, leveraging the strengths of each. In our methodology, we began by designing a physics-guided Fourier-featured DeepONet, drawing inspiration from the intrinsic physical behavior of the target solution. Subsequently, we train this network to primarily learn the low-fidelity solution, utilizing an extensive dataset. This process ensures a comprehensive grasp of the foundational solution patterns. Following this foundational learning, the low-fidelity deep operator network’s output is enhanced using a physics-guided Fourier-featured residual deep operator network. This network refines the initial low-fidelity output, achieving the high-fidelity solution by employing a small high-fidelity dataset for training. Notably, in our framework, we employ the Fourier feature network as the Trunk network for the DeepONets, given its proficiency in capturing and learning the oscillatory nature of the target solution with high precision. We validate our approach using a well-known 2D benchmark cylinder problem, which aims to predict the time trajectories of lift and drag coefficients. The results highlight that the physics-guided Fourier-featured deep operator network, serving as a foundational building block of our framework, possesses superior predictive capability for the lift and drag coefficients compared to its data-driven counterparts. The bi-fidelity learning framework, built upon the physics-guided Fourier-featured deep operator, accurately forecasts the time trajectories of lift and drag coefficients. A thorough evaluation of the proposed bi-fidelity framework confirms that our approach closely matches the high-fidelity solution with an error rate under $2\%$. This confirms the effectiveness and reliability of our framework, particularly given the limited high-fidelity dataset used during training.

\end{abstract}
\end{frontmatter}
\section{Introduction} \label{sec:Introduction}
Deep neural networks have emerged as a fast and reliable alternative for solving many scientific problems \cite{Goodfellow-et-al-2016}. Recent advancements in hardware and software have made it possible to train extensive neural networks that accurately estimate and predict complex engineering and physical systems. However, constructing these models requires a large number of data samples for training. Due to the costs associated with experiments and computations, most studies only include a limited number of high-fidelity data samples to describe a physical problem \cite{HAN19891619, Paniagua2004578, Sahin202007, Sahin202107}. As a result, training neural networks with only high-fidelity data becomes prohibitively expensive and may lead to inaccurate results.

One possible solution for reducing the high cost of producing high-fidelity data is to incorporate data from models or experiments with varying accuracy into the neural network training procedure using multi-fidelity approaches. Instead of immediately resorting to expensive experimental setups or high-dimensional, high-resolution meshes to obtain accurate high-fidelity data, less accurate datasets can be acquired using prediction methods or coarse meshes to construct an initial low-fidelity surrogate. As expected, the accuracy of these initial results is impractical for subsequent steps and needs to be improved. To mitigate this issue and ensure accuracy, multi-fidelity techniques can be employed to bridge the gap between low-fidelity and high-fidelity spaces. This approach can offer cost-effective and precise results for various problems, such as parametric optimization, which relies on accurate data to provide optimal solutions.

In the context of multi-fidelity model optimization, various methods exist to obtain a low-fidelity model, including simplified physics models, coarser meshes in high-fidelity models, or relaxed solver convergence criteria. However, it is crucial for surrogate models to represent the high-fidelity model accurately. To achieve this, corrections are applied to the low-fidelity models. For instance, techniques like space mapping, shape-preserving, and adaptive response correction are utilized in aerodynamic models. Compared to function approximation surrogates, the main advantage of multi-fidelity model optimization is that it requires fewer high-fidelity data points to construct a physics-based model with the desired accuracy level. This reduction in data dependency improves the efficiency of the optimization algorithm, making the process faster and more cost-effective while maintaining acceptable accuracy levels, \cite{LEIFSSON201098}. For example, in \cite{LEIFSSON201098}, the low-fidelity surrogate, constructed from the datasets obtained from the transonic small-disturbance equation, was corrected using the shape-preserving response correction methodology. They were able to reduce computational and optimization costs by reducing the needed high-fidelity dataset by about 90$\%$. 

Multi-fidelity surrogates have shown promising results for several challenging problems and test conditions. For example, in \cite{LEIFSSON201545}, drag and lift coefficients were optimized for an airfoil under transonic conditions using a multi-fidelity model. The surrogate was constructed using coarse and fine meshes to collect low- and high-fidelity datasets. The authors suggested a multi-level method as a computationally effective solution by performing mapping algorithms such as multi-level optimization, space mapping, and shape-preserving response prediction techniques. Similarly, in \cite{SEN2018434}, a multi-fidelity method was considered as an alternative cheap closure model for evaluating drag in shock-particle interactions, rather than using expensive fine-resolution computational solutions. The authors applied space mapping, radial basis functions, and modified Bayesian kriging as correcting techniques on low-fidelity surrogates to determine the best and most inexpensive closure model.

In \cite{DU2019371}, an accelerated optimization process with the multi-fidelity method was outlined for aerodynamic inverse design by comparing the manifold mapping-based design, which needs less than 20 high-fidelity and 1000 to 2000 low-fidelity evaluations, rather than direct aerodynamic inverse design subject to pattern search with 700 to 1200 high-fidelity model evaluations. Furthermore, in \cite{Ertan2022}, by comparing single and multi-fidelity methods to decrease the drag coefficient of a missile, a significant reduction in optimization time was observed. The authors used a faster semi-empirical missile tool to obtain low-fidelity data.

Recognizing the potential of multi-fidelity surrogates, efforts have been made to reduce the need for high-fidelity data and achieve cost savings. One approach is to enhance mapping methods. For example, in \cite{Robinson20067114}, various mapping approaches were explored to determine surrogate-based optimization schemes that can effectively handle constraints in variable-complexity design problems. They observed a 53\% reduction in calling high-fidelity functions in the case of the wing design problem, resulting in reduced computational costs. In \cite{HUANG2013279}, the co-kriging method was used to create a genetic-based surrogate model for a similar purpose. The genetic-based surrogate model was found to have better accuracy than the low-fidelity model-based surrogate models. Moreover, using the genetic-based model reduced computational costs due to fewer high-fidelity data requirements. To generalize the model, \cite{Daniel2008} focused on the residual while constructing the surrogate model for robust optimization. They created a surrogate model to eliminate errors due to the choice of turbulence model on the accuracy and cost of a numerical solution and applied the model to optimize diffuser geometry. In \cite{HAN2013177}, the accuracy of the available multi-fidelity model was improved by combining gradient-enhanced kriging and generalized hybrid bridge function while constructing the multi-fidelity model, and promising results of the robustness were reported.

To improve the accuracy of multi-fidelity surrogates and reduce the cost of high-fidelity data requirements, deep learning methods have been applied to create low-fidelity surrogates and map low- to high-fidelity spaces. The concept of an intelligent teacher-student interaction in machine learning, where student learning is accelerated with privileged information and corrected by transferring knowledge from teacher to student, is described in \cite{Vapnik201516}. In \cite{Dehghani2017}, fidelity-weighted learning was introduced as a new student-teacher structure for deep neural networks. This approach was evaluated in natural language and information retrieval processing and outlined fast and reliable mapping for weakly- and strongly-labeled data. Inductive transfer and bi-fidelity weighted learning methods were utilized in \cite{De2020543} for uncertainty propagation by constructing neural network surrogates from low- and high-fidelity datasets. Two approaches, partial adaption, and shallow network, were used to map between low- and high-fidelity data. The bi-fidelity weighted method showed promising performance on validation errors for three multidisciplinary examples. 

In \cite{Lu20223210}, a novel approach called the deep operator network (DeepONet) integrated a multi-fidelity neural network model to reduce the desired high-fidelity data and attained an error one order of magnitude smaller. This approach was implemented to compute the Boltzmann transport equation (BTE) and proposed a fast solver for the inverse design of BTE problems. To reduce computational costs for complex physical problems involving parametric uncertainty and partial unknowns, a bi-fidelity modeling approach utilizing a deep operator network was introduced in \cite{De20231432}. This approach was applied to three problems: a nonlinear oscillator, heat transfer, and wind farm system. The evaluation demonstrates that the proposed method significantly improves validation error by increasing its efficiency. In \cite{moya2023approximating}, the authors improved their proposed non-autonomous DeepONet-based framework~\cite{lin2023learning} by incorporating a residual learning approach. This approach merges information from pre-existing mathematical models, enabling precise and efficient predictions in challenging power engineering problems. Finally, in~\cite{zhang2023bayesian}, the authors proposed a deep operator learning framework for computing fine-scale solutions of multiscale Partial Differential Equations (PDEs). Specifically, they trained multi-fidelity homogenization maps using mathematically motivated neural operators.

The objective of this paper is to significantly lower the computational expenses involved in simulating the time-dependent evolution of lift and drag coefficients, without compromising on result resolution. To achieve this, we design a novel physics-guided, bi-fidelity Fourier-featured deep operator learning-based framework, which is constructed using coarse and fine datasets obtained through numerical simulation. To attain the aforementioned objective, we make the following contributions. 
\begin{itemize}
    \item We develop a physics-guided, bi-fidelity  Fourier-featured deep operator learning-based framework (see Section~\ref{subsec:bi-fidelity}). This framework takes an arbitrary undisturbed free-stream velocity as input and produces continuous, oscillatory time trajectories of the lift and drag coefficients for a cylinder within a channel. Our approach begins with the design and training of a physics-guided low-fidelity deep operator network using an extensive dataset that captures the foundational solution patterns. Subsequently, the low-fidelity deep operator network's predictions are enhanced through a physics-guided residual deep operator network. This elevation process transitions the low-fidelity solution to a high-fidelity solution utilizing a small high-fidelity dataset. The use of the developed framework enables a comprehensive analysis of the target solution's time evolution for any given undisturbed free-stream velocity, eliminating the need for time-consuming numerical computations and simulations.
    
    \item  The aforementioned deep operator learning-based framework is constructed using a novel physics-guided approach. This approach utilizes the oscillatory nature of the time trajectory of drag and lift coefficients to transform the problem into a functional inverse problem, thereby reducing the solution space for training the deep operator networks. In this paper, we compare the physics-guided approach with the traditional data-driven approach.
    \item Within this framework, we incorporate the Fourier-featured network as the Trunk network of the DeepONets, thereby leading to the development of Fourier-featured DeepONet. This incorporation harnesses the intrinsic capabilities of the Fourier-featured network, particularly its proficiency for capturing the fluctuations of lift and drag time trajectories. Consequently, Fourier-featured DeepONet demonstrates superior performance compared to Vanilla DeepONet, which often struggles with understanding and precisely modeling these oscillatory patterns.
\end{itemize}

The rest of this paper is organized as follows. Section~\ref{sec:materials-and-methods} outlines the numerical method used to generate data on the lift and drag coefficient trajectories, including the specifics of the simulation setup and parameters. Subsequently, we describe the physics-guided, bi-fidelity Fourier-featured deep operator learning methodology for developing a framework to learn target solutions. Section~\ref{sec:results} presents numerical experiments that evaluate both the low-fidelity and the proposed bi-fidelity deep operator learning framework using their respective datasets. Section~\ref{sec:discussion} discusses the results and their implications in detail. Finally, in the concluding Section~\ref{sec:Conclusion}, we encapsulate our key findings and conclusions. 

\section{Methodology and Description of the Tools} \label{sec:materials-and-methods}
This paper proposes a novel approach called Physics-Guided Bi-Fidelity Fourier-Featured Deep Operator Learning to address the challenge of minimizing the cost associated with acquiring high-fidelity data. The approach is applied specifically to accurately estimate the fluctuations of the drag and lift coefficients over time for a cylinder in a channel operating under low Reynolds number conditions.

In order to obtain the necessary data, both low-fidelity and high-fidelity data are collected by simulating the flow around the cylinder using a commercial solver, Ansys Fluent. The data collection process involves varying the Reynolds number within an interval centered around 100. Coarse (low-fidelity) and fine (high-fidelity) meshes are used to capture the flow characteristics. Low-fidelity data is employed to establish an initial deep operator learning network that accepts velocity and time as inputs. To refine the predictions of the low-fidelity deep operator network, a secondary deep operator learning network utilizing high-fidelity data has been integrated. This integration leads to a more precise prediction of time trajectories for drag and lift coefficients. Please note that this study employs velocity as an input parameter. Recognizing the impact of inlet velocity on the pressure field, our future research will leverage detailed pressure data surrounding the cylinder as an input parameter to enhance further understanding.

In the following sections of this paper, we provide detailed descriptions of the numerical method used for data collection. This includes the simulation setup and parameters. Additionally, we delve into the bi-fidelity Fourier-featured deep operator learning framework, explaining its architectural components and training process. These advancements together contribute to a more efficient and cost-effective approach for estimating fluctuations in the drag and lift coefficients over time of the cylinder in the channel under low Reynolds number conditions.

\subsection{Numerical Approach} \label{subsec:Cylinder-CFD}
\begin{figure}[!t] 
\centering
\captionsetup{justification=centering,margin=1pt}
    \begin{tikzpicture}[scale=5]	
        \draw [line width=1pt] (0,0)-- (0,0.41);
        \draw [line width=1pt] (0,0.41)-- (2.2,0.41);
        \draw [line width=1pt] (2.2,0.41)-- (2.2,0);
        \draw [line width=1pt] (2.2,0)-- (0,0);
        \draw[line width=0.75pt]  (0.2,0.2) circle (0.05);
        \draw [fill=black] (2.2,0.41) circle (0.3pt);
        \node[text width=35, align=center, font=\footnotesize,] at (2.06, 0.35) {$2.2,$ $0.41$};
        \draw [fill=black] (0,0.0) circle (0.3pt);
        \node[text width=15, align=center, font=\footnotesize,] at (0.06, 0.04) {$0,$ $0$};
        \draw [fill=black] (0.2,0.2) circle (0.1pt);	
        \node[text width=35.0, align=right, font=\footnotesize, rotate=0] at (0.37, 0.2) {$D = 0.1$ (0.2, 0.2)};	
        \draw [->,line width=0.75pt] (0,0)-- (0,0.15);
        \draw [->,line width=0.75pt] (0,0)-- (0.15,0);
        \draw (0.1,-0.017) node[anchor=north west] {\textbf{x}};
        \draw (-0.1,0.17) node[anchor=north west] {\textbf{y}};
        \draw [->,line width=1.1pt,color=blue] (-0.1,0.205)-- (0.1,0.205);
        \node[text width=5.0] at (-0.06, 0.25) {\textbf{$\bar{u}$}};
        \node[text width=35, align=center, font=\footnotesize,] at (1.05, 0.35) {No slip};
        \node[text width=35, align=center, font=\footnotesize,] at (1.05, 0.05) {No slip};
        \node[text width=35, align=center, font=\footnotesize,] at (2.05, 0.2) {Pressure outlet};
        \draw [<->,line width=1pt,color=black] (0,0.45)-- (2.2,0.45);
        \node[text width=4.0, fill=white, align=center,] at (1.1, 0.46) {\textbf{L}};
        \draw [<->,line width=1pt,color=black] (2.25,0.0)-- (2.25,0.41);
        \node[text width=4.0, fill=white, align=center,] at (2.245, 0.205) {\textbf{H}};
    \end{tikzpicture}
\caption{2D Benchmark cylinder configuration adapted from \cite{Schafer1996}}
\label{Benchmark}	
\end{figure}
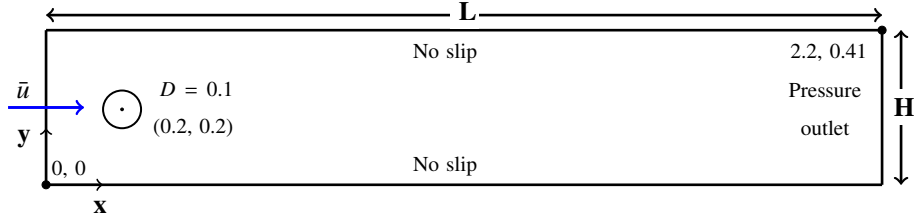

We consider a cylinder with a diameter of $D=0.1m$ placed inside a rectangular channel of length $L=2.2m$ and height $H=0.41m$, as illustrated in Figure~\ref{Benchmark} (replicated from \cite{Schafer1996}). This benchmark simulation investigates two- and three-dimensional laminar flows around a cylinder and has been studied by several research groups using various numerical approaches. The simulation is detailed in terms of drag, lift, and Strouhal number. The $x$-axis (streamwise direction) and $y$-axis are aligned along the length and height of the channel, respectively. The center of the cylinder is positioned at $(0.2, 0.2)$, slightly off-center along the $y$-axis from the center line of the channel to induce vortex shedding. The left and right boundaries of the channel are taken as the inlet and outlet, respectively. The involved quantities are normalized with the cylinder's diameter, $D$. The flow geometry is illustrated in Figure~\ref{Benchmark}. The top and bottom walls of the channel and cylinder solid surface follow the no-slip boundary condition, whereas the outlet boundary condition is set for the exit. A parabolic velocity condition, defined in equation~\eqref{eq:Uparabolic}, is applied to the inlet boundary.

\begin{equation} \label{eq:Uparabolic}
	u_{in}(y) = u_{\max}\left[1-\left(2\frac{y-H/2}{H}\right)^2\right].
\end{equation}
Here, $u_{in}(y)$ represents the velocity vector, and $u_{\max}$ denotes the maximum velocity at the centerline of the channel. The mean velocity of the flow can be calculated using equation~\eqref{eq:Umean}.
\begin{equation} \label{eq:Umean}
	\bar{u} = \frac{1}{H}\int_{0}^Hu_{in}(y)dy. 
\end{equation}
When $u_{\max}$ is set to $1.5 m/s$, the mean velocity around the cylinder is $\bar{u}=1m/s$. This results in a Reynolds number of 100, based on Equation~\eqref{eq:Re}, where $\nu = 0.001 m^2/s$ is the kinematic viscosity of the fluid.
\begin{equation} \label{eq:Re}
Re = \frac{\bar{u}D}{\nu}.
\end{equation}

To create the low- and high-fidelity datasets, the Reynolds number was adjusted to values between 90 and 110 by varying the maximum velocity. For each inlet condition, the lift $F_{L}$ and drag $F_{D}$ were instantaneously calculated on the cylinder surface using equation~\eqref{eq:F}. Here, $S$ denotes the cylinder surface, and $n_x$ and $n_y$ represent the $x$ and $y$ components of the normal vector on $S$, respectively. The tangential velocity on the cylinder surface, $u_{tg}$, is obtained using a tangent vector of $tg = (n_x, -n_y)$.

\begin{equation} \label{eq:F}
    F_D = \int_{S}(\rho\nu\frac{\partial u_{tg}}{\partial n}{n_y}-P{n_x})dS,  \hspace{10pt} F_L = -\int_{S}(\rho\nu\frac{\partial u_{tg}}{\partial n}{n_x}+P{n_y})dS
\end{equation}
The lift $C_{L}$ and drag $C_{D}$ coefficients can be estimated using the corresponding forces on the cylinder surface, as shown in equation \eqref{eq:CDL}.
\begin{equation} \label{eq:CDL}
	C = \frac{2F}{\rho\bar{u}^2D}.
\end{equation}

The Fluent laminar solver is used for numerical simulations with transient settings. Unstructured grids are used for the simulations, with refinement near the cylinder wall. A coarse mesh containing around 3,000 cells is used to obtain low-fidelity data, while high-fidelity data is achieved with a refined mesh of around 87,700 cells. The Courant (CFL) number is kept below 0.9 for each mesh and inlet velocity condition. Additionally, the convergence criteria for the inner iterations is set to $10^{-8}$.

Figure~\ref{DragLift_ref} compares the lift coefficient ($C_{L}$) and drag coefficient ($C_{D}$) of the reference \cite{Schafer1996} with the fine and coarse mesh results. It can be observed that by using the fine mesh results, the maximum values of $C_{L}$ and $C_{D}$ overlap with the reference data, thus validating the results.

\usetikzlibrary {spy}
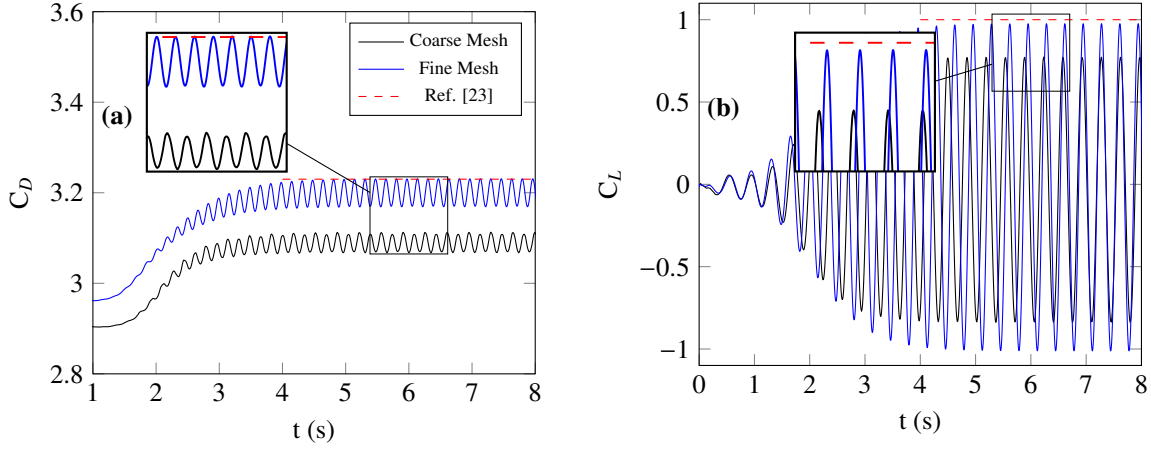
\begin{figure}[!t] 
\captionsetup{justification=centering}
    \begin{subfigure}[!bl]{0.45\textwidth}
    \captionsetup{justification=centering}
    \begin{tikzpicture}[spy using outlines={magnification=1.8, size=1.84cm, connect spies}]	
        \begin{axis}[width=\linewidth, xlabel={t (s)}, ylabel={C$_D$}, xmin = 1.0, xmax=8, ymin=2.8, ymax=3.6, xtick={0,1,...,10}, ytick={0,0.2,...,6}, legend pos=north east,]
        \addplot [smooth, black] table[x index=0, y index=1, each nth point=20]{data/CoarsedragRe100.txt}; 	
        \label{CoarseMeshDrag}; \addlegendentry[font=\scriptsize,]{Coarse Mesh};
        \addplot [smooth, blue] table[x index=0, y index=1, each nth point=20]{data/FineDragRe100.txt}; 	
        \label{FineMeshDrag}; \addlegendentry[font=\scriptsize,]{Fine Mesh};
        \addplot [dashed, red] table{data/RefDragRe100.txt}; 	         
        \label{RefDrag}; \addlegendentry[font=\scriptsize,]{Ref. \cite{Schafer1996}};
        \coordinate (spypoint) at (axis cs:6.0, 3.15);
        \coordinate (magnifyglass) at (axis cs:2.96, 3.4);
        \end{axis}
        \spy on (spypoint) in node [fill=white] at (magnifyglass);
        \node at (0.32, 3.4) {\textbf{(a)}};
        \end{tikzpicture}
    \end{subfigure}
\hspace{5pt}
    \begin{subfigure}[!br]{0.45\textwidth}
    \captionsetup{justification=centering}
        \begin{tikzpicture}[spy using outlines={magnification=1.8, size=1.84cm, connect spies}] 
        \begin{axis}[width=\linewidth, xlabel={t (s)}, ylabel={C$_L$}, xmin = 0.0, xmax=8, ymin=-1.1, ymax=1.1, xtick={0,1,...,10}, ytick={-1,-0.5,...,1}, legend pos=north east,]
        \addplot [smooth, black] table[x index=0, y index=1, each nth point=20]{data/CoarseliftRe100.txt}; 	
        \label{CoarseMeshLift}; 
        \addplot [smooth, blue] table[x index=0, y index=1, each nth point=20]{data/FineLiftRe100.txt};
        \label{FineMeshLift}; 
        \addplot [dashed, red] table{data/RefLiftRe100.txt}; 	             
        \label{RefLift};
        \coordinate (spypoint) at (axis cs:6.0, 0.8);
        \coordinate (magnifyglass) at (axis cs:3, 0.5);
        \end{axis}
        \spy on (spypoint) in node [fill=white] at (magnifyglass);
        \node at (0.32, 3.4) {\textbf{(b)}};
        \end{tikzpicture}
    \end{subfigure}
\caption{Comparison of low (coarse mesh) and high-fidelity (fine mesh) $C_D$ and $C_L$ with the reference \cite{Schafer1996} at $Re=100$: a) Drag Coefficient, $C_D$, and b) Lift Coefficient, $C_L$.}
\label{DragLift_ref}
\end{figure}

\subsection{Bi-Fidelity Fourier-Featured Deep Operator Learning} \label{subsec:bi-fidelity}
This section describes the proposed bi-fidelity Fourier-featured deep operator learning framework for approximating the operators that predict the drag and lift coefficients over time.
\subsubsection{Deep Operator Learning} \label{subsec:DeepONet}
This paper proposes a bi-fidelity Fourier-featured deep operator learning framework to approximate the operator mapping~$\mathcal{G}$ between the undisturbed free-stream velocity in the channel, denoted as $\overline{u} \in \mathcal{U}$, and the time-dependent trajectories of the drag and lift coefficients of the cylinder, denoted as $C_D$ and $C_L$, respectively, for $t \in [t_0,t_f]$, \ie
\begin{equation}\label{eq:operator}
\mathcal{G}: \overline{u} \mapsto v
\end{equation}
where $v \in \{C_D, \ C_L\}$. In the above, the input domain for the undisturbed free-stream velocity $\overline{u}$ is given by $\mathcal{U} = [0.9, 1.1]$ m/s, and the time domain for the coefficients is $[t_0, t_f] = [0, 8]$ seconds. 

To approximate the operator $\mathcal{G}$, we will design a bi-fidelity Fourier-featured deep operator learning framework, denoted $\mathcal{G}^f_\theta$, where $\theta$ is the vector of trainable parameters. This framework consists of two deep operator networks, which we will introduce in the next sections: a low-fidelity deep operator network, denoted $\mathcal{G}^c_{\theta^c}$, and a residual deep operator network, denoted $\mathcal{G}^\epsilon_{\theta^{\epsilon}}$. The proposed framework then satisfies:
\begin{equation}\label{eq:fine-operator}
\mathcal{G}^f_\theta = \mathcal{G}^c_{\theta^c} + \mathcal{G}^\epsilon_{\theta^\epsilon} \approx \mathcal{G}
\end{equation}

\begin{figure}[h]
    \centering
    \includegraphics[width=1\textwidth]{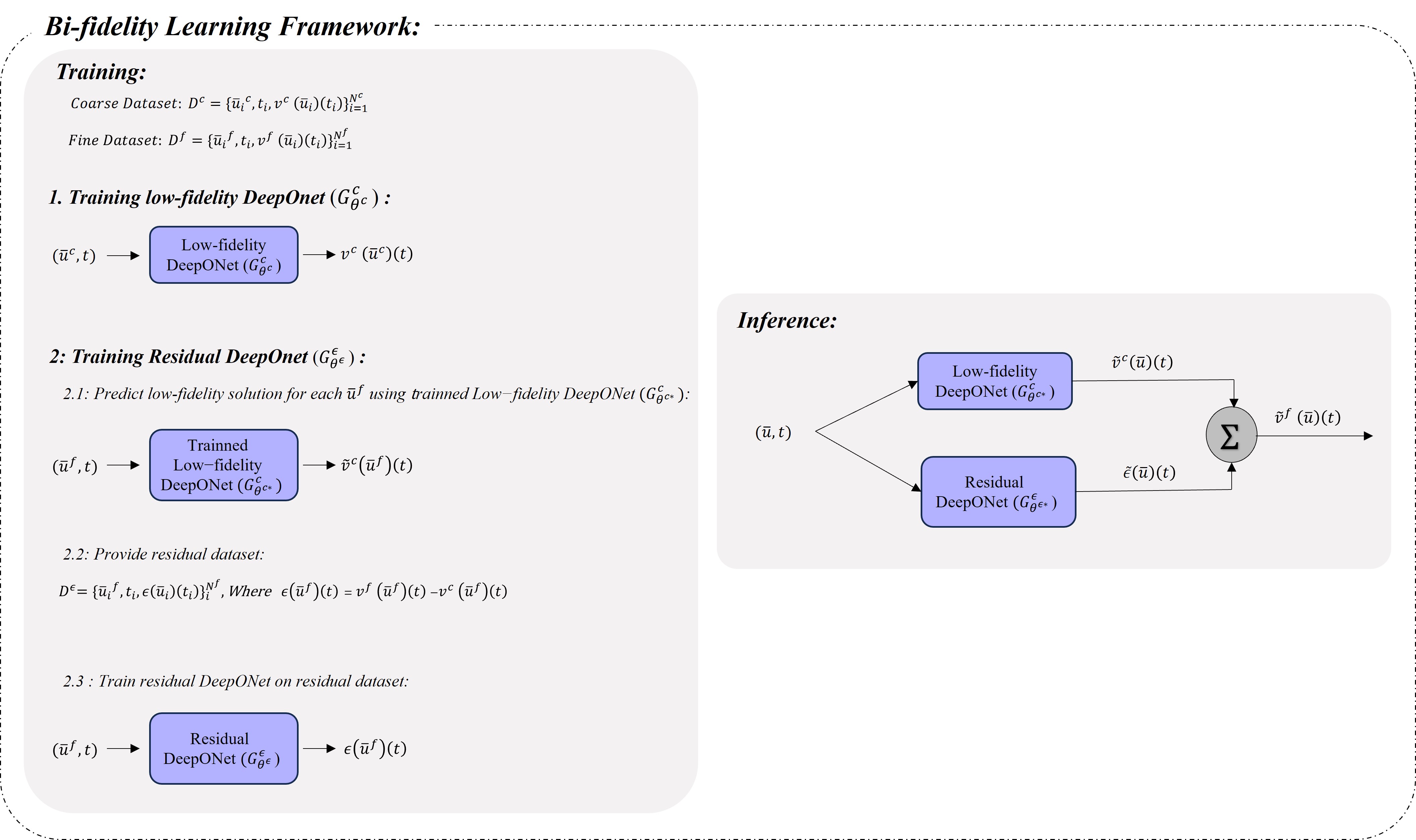}
    \caption{Training and inference phases of the developed bi-fidelity learning framework. In this figure, \(\overline{u}\) represents the undisturbed free-stream velocity and \(t\) denotes time, both serving as inputs to the model. The variable $v \in \{C_D, C_L$\}, indicates the target solution. The superscripts \(c\) and \(f\) correspond to coarse and fine grades of the target solution, respectively, while the ($\sim$) symbolizes the approximate solution. The term $\epsilon\ = v^f-v^c$ is defined as residuals which is the difference between the high-fidelity (\(v^f\)) and the low-fidelity (\(v^c\)) solutions. Notably, within this framework, both the low and residual DeepONets can be data-driven or physics-guided deep operators. Their specific types are not depicted in the figure to highlight this flexibility.}
    \label{fig:Schematic_Bi}
\end{figure}

\begin{figure}[h]
    \centering
    \includegraphics[width=1\textwidth]{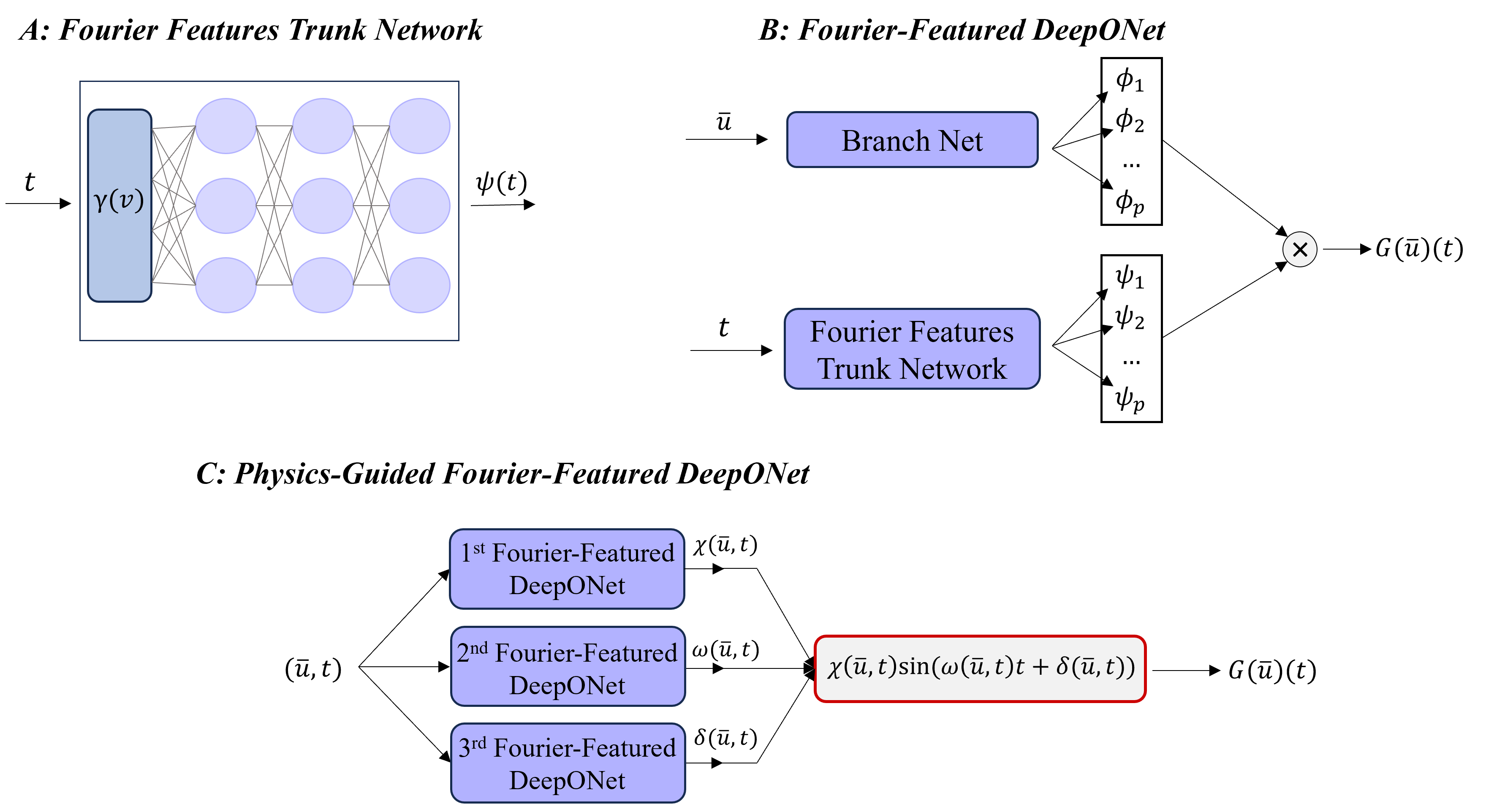}
    \caption{Hierarchy of the implemented models. 
(A) Fourier Features Trunk Network. 
(B) Fourier-featured DeepONet structure. 
(C) physics-guided Fourier-featured DeepONet model: Integrating multiple DeepONets through physical equations to achieve the target solution. In this figure, \(\overline{u}\) represents the undisturbed free-stream velocity, and \(t\) denotes time, both serving as inputs to the model. Additionally, \(\phi\) and \(\psi\) are the basis for the branch and trunk networks, respectively. }
    \label{fig:Schematic_models}
\end{figure}

\subsubsection{Low-Fidelity Deep Operator Network} \label{subsubsec:low-fidelity-DeepONet}
Using a coarse mesh in simulations of complex systems is an effective approach to reduce the computational demands of the simulation process. This methodology simplifies the system's complexity while generating abundant data that will significantly enhance the training process for deep operator networks. However, the adoption of a coarse mesh also introduces its own set of challenges. Notably, it omits fine-grained details in the system's dynamics, leading to approximation errors. These errors may initially be minor but can compound over time, potentially resulting in significant discrepancies and impacting the accuracy and reliability of the deep operator network’s predictions.

The proposed bi-fidelity framework capitalizes on the maximum advantages of the low-fidelity deep operator network while deploying a residual deep operator network to bridge the gap between the coarse and fine operators. This strategy enables the framework to benefit from the computational cost-effectiveness of the low-fidelity deep operator network without compromising the high accuracy characteristic of the fine operator.

The proposed bi-fidelity learning approach first designs a low-fidelity deep operator network with trainable parameters $\theta^c$, denoted as $\mathcal{G}^c_{\theta^c}: \overline{u} \mapsto v^c$. This deep operator network maps the undisturbed free-stream velocity $\overline{u} \in \mathcal{U}$ onto the low-fidelity solution $v^c$ obtained using coarse mesh simulation. The target solution for the low-fidelity deep operator network, $v^c \in \{C_D, C_L\}$, corresponds to temporal trajectories of either the drag or lift coefficients within the specified time domain $[t_0,t_f]$. We will design this low-fidelity deep operator network to be either a physics-guided deep operator network (see Section~\ref{subsubsec: Physics-Guided-DeepONet}) or a data-driven deep operator network with Fourier-feature layers (see Section~\ref{subsubsec: DeepONet}).

\textit{Training the low-fidelity deep operator network.} To train the coarse deep operator network, we minimize the following loss function:
\begin{equation}
\mathcal{L}(\theta^c;\mathcal{D}^c) =\frac{1}{N} \sum_{j=1}^N 
|v^c_j(\overline{u}_j)(t_j) - \mathcal{G}^c_{\theta^c}(\overline{u}_j)(t_j)|^2
\end{equation}
using the dataset of $N$ triplets: $\mathcal{D}^c = \{u_i, t_i, v^c(\overline{u}_i)(t_i) \}_{i=1}^{N}$, where $v \in \{C_D, C_L\}$ for a given undisturbed free-stream velocity~$\overline{u}_i \in \mathcal{U}$. In practice, for each velocity $\overline{u}_i$, we can obtain $q$ samples of the target solution by evaluating it at $q$ different times $\{t_i^j\}_{j=1}^q$.

Since generating the low-fidelity dataset is computationally less expensive, we assume that this dataset can be large enough. This large dataset is particularly advantageous for training our low-fidelity deep operator network as it provides a diverse range of examples to learn from, thereby enhancing the efficacy and generalization ability of the model.

Furthermore, we note that a trainable low-fidelity deep operator network could be replaced by either an efficient simulator or a Physics-Informed Neural Network (PINN)~\cite{karniadakis2021physics}. An efficient simulator could streamline the computation process by providing faster approximations, while a PINN could potentially offer a more efficient model by encoding the simulator's equations directly into the loss function.

\subsubsection{Residual Deep Operator Network} \label{subsubsec:high-fidelity-DeepONet}
To approximate the fine or high-fidelity operator, we propose a residual deep operator learning strategy that uses the output from the trained low-fidelity deep operator network $\mathcal{G}^c_{\theta^{c*}}$. Figure~\ref{fig:Schematic_Bi} provides a visual representation of the proposed framework. The framework involves defining the residual operator: 
\begin{equation}
\mathcal{G}^\epsilon:=\mathcal{G} - \mathcal{G}^c_{\theta^{c*}}
\end{equation}
which estimates the difference between the true fine operator and the output of the trained low-fidelity deep operator network.

We approximate the residual operator mentioned above using a second deep operator network. Specifically, the proposed residual deep operator network, denoted by $\mathcal{G}^\epsilon_{\theta^\epsilon}$, maps the undisturbed free-stream velocity $\overline{u} \in \mathcal{U}$ to the residual solution $\epsilon = v^f-v^c$, where $v^f$ and $v^c$ are the fine solution and the predicted coarse solution, respectively. Note that this second network fine-tunes the predictions generated by the low-fidelity deep operator network, facilitating precise adjustments to the high-fidelity solution.

\textit{Training the residual deep operator network.} To train the residual deep operator network, we minimize the following loss function:
\begin{equation}
\mathcal{L}(\theta;\mathcal{D}^\epsilon) =\frac{1}{N} \sum_{j=1}^N |\epsilon(\overline{u}_j)(t_j) - \mathcal{G}^{\epsilon}_{\theta^{\epsilon}}(\overline{u}_j)(t_j)|^2,
\end{equation}
using the dataset of $N$ triplets $\mathcal{D}^\epsilon = \{\overline{u}_i, t_i, \epsilon_j(\overline{u}_j)(t_j) \}_{i=1}^{N}$, where $\epsilon_j(\overline{u}_j)(t_j) =v^{f}(\overline{u}_j)(t_j)-\mathcal{G}^{c}_{\theta^{c*}}(\overline{u}_j)(t_j)$ denotes the residual error, \ie the difference between the high-fidelity solution, obtained via fine-mesh simulation, and the corresponding prediction generated by the trained low-fidelity deep operator network.

Obtaining high-fidelity solutions for the dataset $\mathcal{D}^\epsilon$ can be expensive, resulting in a smaller dataset size compared to the coarse dataset $\mathcal{D}^c$. This expense is due to the substantial computational resources required for high-resolution CFD simulations. While we used a uniform random sampling method to collect data, it may not be ideal for costly high-fidelity data. Adopting a more strategic sampling approach, aimed at specific areas of complexity or interest, could potentially enhance the proposed framework’s performance. We will explore this strategic sampling approach in future work.

\textit{Inference.} After training the low-fidelity and residual deep operator networks, we predict the time evolution of lift and drag coefficients for an arbitrary undisturbed free-stream velocity in a channel using a two-step process. First, we obtain a low-fidelity approximation of the solution using the trained low-fidelity deep operator network. This serves as our initial prediction/guess. Then, we use the residual deep operator network to predict the error between the true solution over time and our initial guess. Finally, we compute the approximation of the true solution by adding the outputs from both the trained low-fidelity and residual deep operator networks. Algorithm \ref{alg:bi-fidelity-learning} summarizes the details of this two-step process for predicting the lift and drag coefficients for a given time partition~$\mathcal{P}:=\{t_0, \ldots, t_f$\}.


\begin{algorithm}[t]
\DontPrintSemicolon
\SetAlgoLined
\textbf{Require:} The trained \textit{low-fidelity} deep operator network~$\mathcal{G}^c_{\theta^{c*}}$, the trained \textit{residual} deep operator network~$\mathcal{G}^\epsilon_{\theta^{\epsilon*}}$, the undisturbed free-stream velocity of the fluid in the channel $\overline{u} \in \mathcal{U}$, a given time partition $\mathcal{P}$.

\textbf{Step 1:} Use the trained low-fidelity deep operator network $\mathcal{G}^c_{\theta^{c*}}$
to predict the low-fidelity solution on $\mathcal{P}$ for the given velocity~$\overline{u}$, \ie $\{v^c(\overline{u})(t)\equiv \mathcal{G}^c_{\theta^{c*}}(\overline{u})(t) : t \in \mathcal{P} \}$, where $v^c \in \{ C_D, C_L\}$.

\textbf{Step 2.} Use the trained residual deep operator network~$\mathcal{G}^\epsilon_{\theta^{\epsilon*}}$ to predict the errors~$\{\epsilon(\overline{u})(t) \equiv \mathcal{G}^\epsilon_{\theta^{\epsilon*}}(\overline{u})(t) : t \in \mathcal{P}\}$ between the high-fidelity solution and the predicted low-fidelity solution on~$\mathcal{P}$ for the given velocity~$\overline{u}$.

\textbf{Return.} The predicted high-fidelity solution on~$\mathcal{P}$, \ie $\{v^f(\overline{u})(t):t \in \mathcal{P}\}$, where $v^f(\overline{u})(t) = v^c(\overline{u})(t) +\epsilon(\overline{u})(t)$ and $v^f \in \{ C_D, C_L\}$.
\caption{A Bi-Fidelity Fourier-Featured Deep Operator Framework for Predicting the Lift and Drag Coefficients.}
\label{alg:bi-fidelity-learning}
\end{algorithm}
\subsubsection{Deep Operator Learning} \label{subsubsec: DeepONet}
In this paper, we adopt the Deep Operator Network (DeepONet) proposed in~\cite{2021Lulu1} as the foundational model for constructing our novel, physics-guided, bi-fidelity Fourier-featured deep operator learning-based framework. DeepONet is based on the universal approximation theorem of nonlinear operators~\cite{1995Chen1}. It can approximate any nonlinear continuous operator, which are mappings between infinite dimensional spaces.

Figure~\ref{fig:Schematic_models}B illustrates the DeepONet architecture, which is designed to approximate the target operator using a trainable linear representation. To enable this representation, it is crucial to construct and train two interconnected but distinct sub-neural networks: the Branch network and the Trunk network. The Branch network is primarily responsible for handling the input, denoted as $\overline{u} \in \mathcal{U}$, and produces a vector of trainable basis coefficients, denoted as $\phi$. The Trunk network, on the other hand, decodes the output by processing the output location $t \in T$. Its output is another vector of trainable basis functions, denoted as $\psi$. The desired linear representation is achieved by taking the dot product of the outputs from the Branch and Trunk networks. This effectively combines these outputs in a meaningful way to approximate the target operator as follows:

\begin{equation}
 v(\overline{u},t) \approx \mathcal{G}_{\theta}(\overline{u})(t) = \sum_{k=1}^p \phi_k(\overline{u}) \cdot \psi_k(t)
\end{equation}

\subsubsection{Physics-Guided Deep Operator Learning} \label{subsubsec: Physics-Guided-DeepONet}
The oscillatory behavior of the time trajectory of lift and drag coefficients suggests that a physics-guided approach could effectively decipher these patterns. Physics-guided neural networks have a unique advantage in incorporating physical intuition into deep learning, resulting in accurate and robust predictions. This advantage is especially valuable when dealing with complex phenomena, such as predicting drag and lift coefficients across a wide range of input velocities. By incorporating physical intuition into deep learning, the precision, reliability, and generalizability of the predictions can be improved compared to fully data-driven models. Although data-driven models are powerful, they may fail to capture complex dependencies if they are not adequately represented in the dataset. Physics-guided models address this potential limitation by utilizing known relationships and behaviors.

In this paper, we adopt a physics-guided approach along with Deep Operator Neural Networks to transform the problem into an operator/functional inverse problem as illustrated in Figure~\ref{fig:Schematic_models}C. In particular, the oscillatory nature of the lift and drag coefficients over time, as shown in Figure~\ref{DragLift_ref}, suggests the suitability of sinusoidal family functions for modeling this behavior. Based on this observation, we propose integrating this physics-guided approach into our deep operator learning approach. More specifically, our proposed physics-guided deep operator network is defined as follows:
\begin{equation}
\mathcal{G}^{v}_P (\overline{u},t) = \chi(\overline{u},t) \sin(\omega(\overline{u},t) t + \delta(\overline{u},t)).
\end{equation}
Here, $v \in \{{C_D, C_L}\}$ represents either the lift or drag coefficients, while $\chi$, $\omega$, and $\delta$ are the outputs of three distinct DeepONets. These networks take the undisturbed free-stream velocity and time as inputs. In this configuration, $\chi$ learns the amplitude variations over time of the target solution,  $\omega$ understands the frequency of the solution's trajectory, and $\delta$ captures the phase angle. This division of tasks allows our model to capture both the scale and periodicity of the system dynamics. By embedding the physics of the problem into our model, we anticipate enhancing the model's robustness and reliability.

\textit{Training the Physics-Guided Deep Operator Network.} To train the low-fidelity physics-guided deep operator network, we minimize the subsequent loss function:
\begin{equation}
\mathcal{L}(\theta^c;\mathcal{D}^c) =\frac{1}{N} \sum_{j=1}^N 
|v^c_j(\overline{u}_j)(t_j) - \mathcal{G}^c_{P,\theta^c}(\overline{u}_j)(t_j)|^2
\end{equation}
utilizing a dataset comprising \( N \) triplets, denoted as \( \mathcal{D}^c = \{u_i, t_i, v^c(\overline{u}_i)(t_i) \}_{i=1}^{N} \), where $v \in \{C_D, C_L\}$ corresponds to a specific channel undisturbed free-stream velocity \( \overline{u}_i \) from the set \( \mathcal{U} \). Furthermore, to train the residual physics-guided deep operator network, we optimize the following loss function:
\begin{equation}
\mathcal{L}(\theta;\mathcal{D}^\epsilon) =\frac{1}{N} \sum_{j=1}^N |\epsilon(\overline{u}_j)(t_j) - \mathcal{G}^{\epsilon}_{P, \theta^{\epsilon}}(\overline{u}_j)(t_j)|^2,
\end{equation}
using the dataset of $N$ triplets $\mathcal{D}^\epsilon = \{\overline{u}_i, t_i, \epsilon_j(\overline{u}_j)(t_j) \}_{i=1}^{N}$,  the term \(\epsilon_j(\overline{u}_j)(t_j) = v^{f}(\overline{u}_j)(t_j)-\mathcal{G}^{c}_{P,\theta^{c*}}(\overline{u}_j)(t_j)\) represents the residual error.

\textit{Inference.} For a comprehensive understanding, Algorithm~\ref{alg:bi-fidelity-learningP} provides a thorough depiction of the inference phase for the bi-fidelity framework, built upon the physics-guided Deep Operator components.

\begin{algorithm}[t]
\DontPrintSemicolon
\SetAlgoLined
\textbf{Require:} The trained \textit{physics-guided low-fidelity} deep operator network~$\mathcal{G}^c_{P,\theta^{c*}}$, the trained \textit{physics-guided residual} deep operator network~$\mathcal{G}^\epsilon_{P,\theta^{\epsilon*}}$, the undisturbed free-stream velocity of the fluid in the channel $\overline{u} \in \mathcal{U}$, a given time partition $\mathcal{P}$.

\textbf{Step 1:} Use the trained physics-guided low-fidelity deep operator network $\mathcal{G}^c_{P,\theta^{c*}}$
to predict the low-fidelity solution on $\mathcal{P}$ for the given velocity~$\overline{u}$, \ie $\{v^c(\overline{u})(t)\equiv \mathcal{G}^c_{P,\theta^{c*}}(\overline{u})(t) : t \in \mathcal{P} \}$, where $v^c \in \{ C_D, C_L\}$.

\textbf{Step 2.} Use the trained physics-guided residual deep operator network~$\mathcal{G}^\epsilon_{P,\theta^{\epsilon*}}$ to predict the errors~$\{\epsilon(\overline{u})(t) \equiv \mathcal{G}^\epsilon_{P,\theta^{\epsilon*}}(\overline{u})(t) : t \in \mathcal{P}\}$ between the high-fidelity solution and the predicted low-fidelity solution on~$\mathcal{P}$ for the given velocity~$\overline{u}$.

\textbf{Return.} The predicted high-fidelity solution on~$\mathcal{P}$, \ie $\{v^f(\overline{u})(t):t \in \mathcal{P}\}$, where $v^f(\overline{u})(t) = v^c(\overline{u})(t) +\epsilon(\overline{u})(t)$ and $v^f \in \{ C_D, C_L\}$.
\caption{A Physics-Guided Bi-Fidelity Fourier-Featured Deep Operator Framework for Predicting the Lift and Drag Coefficients.}
\label{alg:bi-fidelity-learningP}
\end{algorithm}

\subsubsection{Fourier Features Network}
\label{subsubsec:Fourier-features-network}
To design each DeepONet, we employ a feed-forward neural network for the Branch network. However, traditional feed-forward networks have limitations in capturing high-frequency oscillatory patterns, making it difficult to encode the target output solutions, specifically the oscillatory time trajectory of the drag and lift coefficients. To overcome this limitation, we develop the Fourier-featured DeepONet by adopting the Fourier features network as the Trunk network within the DeepONets' architecture. This network is responsible for encoding the output and can effectively encapsulate the target solution. It is selected for its exceptional ability to capture periodic patterns and oscillatory behaviors, transcending the capabilities of conventional feed-forward layers. 

The Fourier features network, as shown in Figure~\ref{fig:Schematic_models}A relies on a random Fourier mapping expressed as $\gamma(v)=[\sin(Bv),\ \cos(Bv)]^T$ \cite{tancik2020fourier}. The matrix $B$ in $R^{m\times d}$ contains values sampled from a Gaussian distribution $N(0,\sigma^2)$. By integrating a random Fourier mapping with a traditional neural network, the Fourier features network effectively boosts learning capabilities by simplifying high-dimensional data. This method is distinctive in its ability to mitigate spectral bias, a common issue that hinders the learning of high-frequency data components. As a result, it significantly enhances performance across various tasks, making it an ideal choice for the Trunk network in the DeepONet model used for predicting oscillatory time trajectories of lift and drag coefficients.

\section{Numerical Model Results} \label{sec:results}
This section presents the training and testing procedures used in the proposed physics-guided bi-fidelity Fourier-featured deep operator learning framework to predict the drag and lift coefficients of a cylinder.
\subsection{Training and testing datasets} \label{subsec:NN-datasets}

To train and evaluate our physics-guided bi-fidelity Fourier-featured deep operator learning framework, two separate datasets are needed: one for low-fidelity data and another for high-fidelity data. The low-fidelity dataset is obtained by simulating with a coarse mesh, while the high-fidelity dataset is generated using a fine mesh. The fine mesh of the high-fidelity dataset produces more accurate results but comes with a significantly higher computational cost. Within each dataset, two-dimensional numerical simulations were conducted for each $u_j \in \mathcal{U}$, where $j = 1,\ldots,\ N$, specifically for the undisturbed free-stream velocity within the channel. To obtain the training targets, we monitored and recorded the lift and drag coefficients of the cylinder over time. The values of these coefficients were collected within the specified time domain. Another key difference between these datasets, apart from the mesh resolution, is their size. The low-fidelity dataset is much larger than its high-fidelity counterpart, primarily due to cost-effectiveness and affordability.

In this paper, we utilized 150 sets of simulation data to train the low-fidelity DeepONet model and 50 sets to train the residual DeepONet. For testing purposes, we allocated 10\% of the data, while the remaining 90\% was used for training. To ensure independence, we employed a simple random selection method to transform the input functions into separate samples. This approach fosters the development of a more generalized model, enhancing its applicability under diverse operating conditions.
\subsection{Neural Networks and Training Protocols} \label{subsec:NN-training}
To ensure the model's effectiveness and establish an efficient training process, we conducted routine hyperparameter optimization. This optimization aimed to identify the optimal architecture and suitable settings for the training process. The results of this optimization are presented in Table~\ref{tab:selected-hps}. For the training of all DeepONets discussed in this paper, physics-guided and data-driven, we employed the settings detailed along with the Adam optimizer, coupled with a "reduce-on-plateau" learning rate scheduler.

\begin{table}[ht]
\caption{Hyper-parameter optimization results}
\centering
\renewcommand{\arraystretch}{1}
\begin{tabular}{ p{5cm} |c }
\hline
\textbf{Hyper-parameter} &  \textbf{Optimum value  }              \\
\hline
Initial learning rate $(\eta)$ &  $1 \times 10^{-4}$          \\
Number of layers in each net &   3 \\
Number of neurons in each layer & 100 \\
Activation function   &     ReLU     \\
Loss function &      MSE  
\label{tab:selected-hps}
\end{tabular}
\end{table}

\subsection{Low-Fidelity Deep Operator Learning 
Results} \label{subsec:low-fidelity-results}
For low-fidelity learning, we designed both physics-guided and data-driven low-fidelity deep operator networks and assessed their performance. After training each operator on the low-fidelity training dataset, we evaluated their effectiveness using a low-fidelity test dataset. This test dataset constitutes 10 percent of the total low-fidelity data, which was excluded from the training phase. It includes trajectories spanning $q= 300$ time steps within the output domain $[t_0,t_f]$. 

To visualize the performance of the proposed low-fidelity models, we randomly selected an undisturbed free-stream velocity \( u_j \in \mathcal{U} \) from the low-fidelity test dataset and predicted the drag and lift coefficients using both physics-guided and data-driven deep operator networks. Figure~\ref{DragLift_lowfidelity
_Physics} and Figure~\ref{DragLift_lowfidelity
_Data} depict these coefficients for the chosen velocity. The displayed results underscore the exceptional predictive prowess of the low-fidelity DeepONet models; their predictions align closely with the true values derived from CFD simulations using a coarse mesh.

\begin{figure}[!h] 
    \begin{subfigure}[!bl]{0.45\textwidth}
    \captionsetup{justification=centering}
        \begin{tikzpicture}[spy using outlines={magnification=2, size=1.6cm, connect spies}]
        \begin{axis}[width=\linewidth, xlabel={$t (s)$}, ylabel={$C_D$}, xmin=1, xmax=8, 
        ymin=2.6, ymax=2.85, 
        xtick={0,1,...,10}, ytick={0,0.05,...,4}, ]           
        \addplot [smooth, red] table{data/LFTrue_drag_Physics.txt}; 
        \label{CdTrue};  \addlegendentry[font=\scriptsize,]{Low-Fidelity Data}	
        \addplot [dashed, black] table{data/LFPredic_drag_Physics.txt}; 
        \label{CdPredic};  \addlegendentry[font=\scriptsize,]{Low-Fidelity DeepONet}
        \coordinate (spypoint) at (axis cs:7.0, 2.755);
        \coordinate (magnifyglass) at (axis cs:4.5, 2.675);
        \end{axis}
        \spy on (spypoint) in node [fill=white] at (magnifyglass);
        \node at (0.5, 3.4) {\textbf{(a)}};
        \end{tikzpicture}
    \end{subfigure}
    \hspace{5pt}
    \begin{subfigure}[!br]{0.45\textwidth}
    \captionsetup{justification=centering}
        \begin{tikzpicture}[spy using outlines={magnification=2, size=1.6cm, connect spies}]
        \begin{axis}[width=\linewidth, xlabel={$t (s)$},ylabel={$C_L$}, xmin=0.0, xmax=8, ymin=-0.8, ymax=0.8, 
        xtick={0,1,...,10}, ytick={-0.6,-0.3,...,1}, ] 
        \addplot [smooth, red] table{data/LFTrue_lift_Physics.txt}; 
        \label{ClTrue};           
        \addplot [dashed, black] table{data/LFPredic_lift_Physics.txt}; 
        \label{ClPredic};          
        \coordinate (spypoint) at (axis cs:7.0, 0.53);
        \coordinate (magnifyglass) at (axis cs:4.5, 0.41);
        \end{axis}
        \spy on (spypoint) in node [fill=white] at (magnifyglass);
        \node at (0.5, 3.4) {\textbf{(b)}};
        \end{tikzpicture}
    \end{subfigure}
\caption{Comparison between the low-fidelity solution (data) time trajectory and the corresponding low-fidelity physics-guided Fourier-featured DeepONet prediction for a random undisturbed free-stream velocity~($\overline{u}=0.94~ m/s$): a) Drag Coefficient, $C_D$, and b) Lift Coefficient, $C_L$.}
\label{DragLift_lowfidelity
_Physics}
\end{figure}
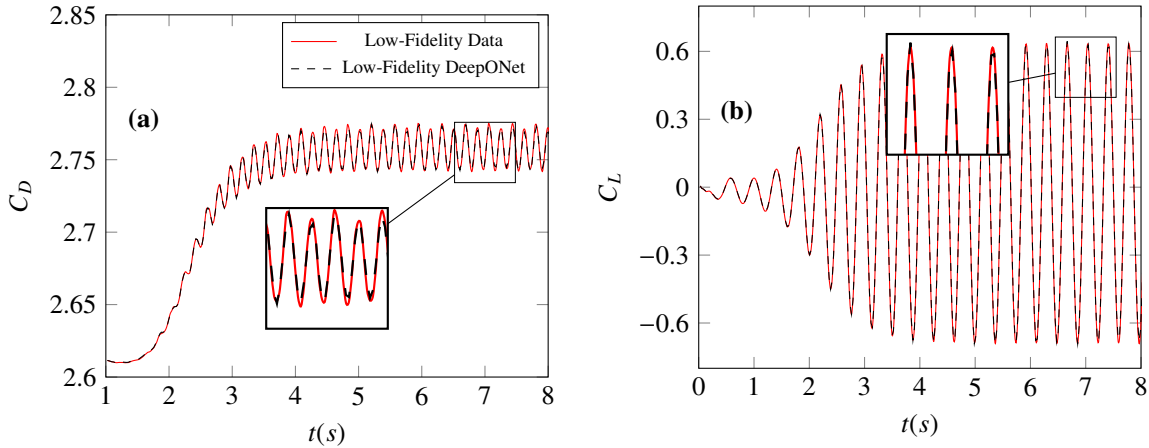

\begin{figure}[!h] 
    \begin{subfigure}[!bl]{0.45\textwidth}
    \captionsetup{justification=centering}
        \begin{tikzpicture}[spy using outlines={magnification=2, size=1.6cm, connect spies}]
        \begin{axis}[width=\linewidth, xlabel={$t (s)$},ylabel={$C_D$}, xmin = 1, xmax=8, ymin=2.6, ymax=2.85, 
        xtick={0,1,...,10}, ytick={0,0.05,...,4}, ]           
        \addplot [smooth, red] table{data/LFTrue_drag_Data.txt}; 
        \label{CdTrue}; \addlegendentry[font=\scriptsize,]{Low-Fidelity Data}	
        \addplot [dashed, black] table{data/LFPredic_drag_Data.txt}; 
        \label{CdPredic}; \addlegendentry[font=\scriptsize,]{Low-Fidelity DeepONet}
        \coordinate (spypoint) at (axis cs:7.0, 2.755);
        \coordinate (magnifyglass) at (axis cs:4.5, 2.675);
        \end{axis}
        \spy on (spypoint) in node [fill=white] at (magnifyglass);
        \node at (0.5, 3.4) {\textbf{(a)}};
        \end{tikzpicture}
    \end{subfigure}
    \hspace{5pt}
    \begin{subfigure}[!br]{0.45\textwidth}
    \captionsetup{justification=centering}
        \begin{tikzpicture}[spy using outlines={magnification=2, size=1.6cm, connect spies}]
        \begin{axis}[width=\linewidth, xlabel={$t (s)$},ylabel={$C_L$}, xmin = 0.0, xmax=8, ymin=-0.8, ymax=0.8, 
        xtick={0,1,...,10}, ytick={-0.6,-0.3,...,1}, ] 
        \addplot [smooth, red] table{data/LFTrue_lift_Data.txt}; 
        \label{ClTrue};         
        \addplot [dashed, black] table{data/LFPredic_lift_Data.txt}; 
        \label{ClPredic};       
        \coordinate (spypoint) at (axis cs:7.0, 0.53);
        \coordinate (magnifyglass) at (axis cs:4.5, 0.41);
        \end{axis}
        \spy on (spypoint) in node [fill=white] at (magnifyglass);
        \node at (0.5, 3.4) {\textbf{(b)}};
        \end{tikzpicture}
    \end{subfigure}
\caption{Comparison between the low-fidelity solution (data) time trajectory and the corresponding low-fidelity data-driven  Fourier-featured DeepONet prediction for a random undisturbed free-stream velocity~($\overline{u}=0.94 ~m/s$): a) Drag Coefficient, $C_D$, and b) Lift Coefficient, $C_L$.}
\label{DragLift_lowfidelity
_Data}
\end{figure}
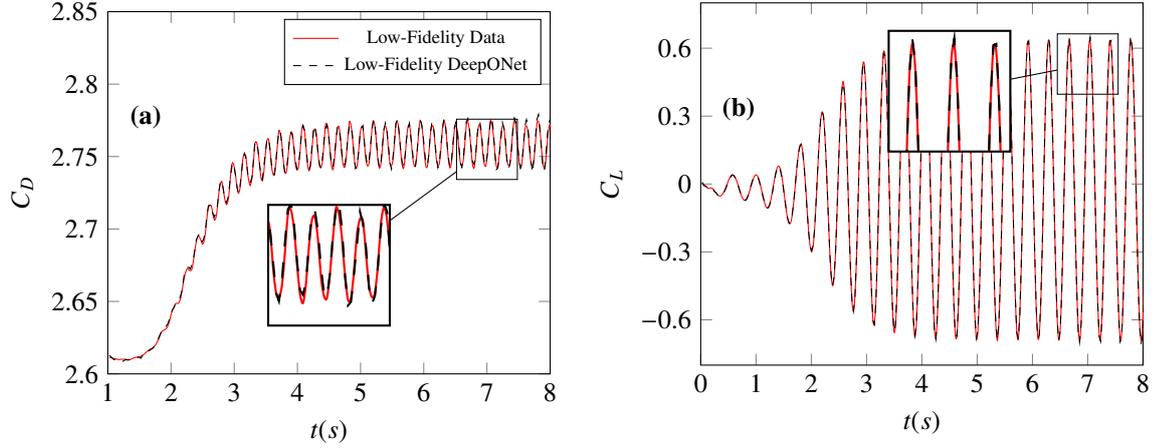

\subsubsection{Comparative Performance Analysis of Low-Fidelity Deep Operator Networks} 
In order to evaluate and compare the generalization capabilities of the proposed physic-guided and data-driven models, we computed the mean and standard deviation of the $L_1$ and $L_2$-relative errors for each model using 15 test trajectories. These were calculated between the actual and forecasted values of the drag and lift coefficients' time trajectories using the low-fidelity testing dataset. As shown in Table \ref{table:DeepONet-resultsLow_Drag} and Table \ref{table:DeepONet-resultsLow_Lift}, the physics-guided low-fidelity DeepONet consistently outperformed its data-driven counterpart in terms of accuracy and reduced errors. A key contributing factor to this enhanced performance is that the physics-guided approach inherently narrows down the solution space, ensuring more precise predictions by leveraging established physical principles and constraints.
However, both the physics-guided and data-driven low-fidelity DeepONets achieved errors below $2\%$ for both coefficients. This demonstrates the effectiveness of the low-fidelity deep operator network in providing accurate predictions within the context of the problem domain, showcasing its potential to serve as the basis for the bi-fidelity framework.

\begin{table}[H]
\caption{Mean and standard deviation of $L_1$ and $L_2$ relative errors between low-fidelity solution (data) and low-fidelity Fourier-featured deep operator networks' predictions for the Drag coefficient, $C_D$.}
\centering

\begin{tabular}{c |  c  c  c  c} 
\hline

\textbf{Low-fidelity model}  &\textbf{mean~  $L_1$}&\textbf{st.dev.~$L_1$}&\textbf{{mean~  $L_2$}}&\textbf{st.dev.~$L_2$} \\
\hline
\hline

\textbf{Physics-guided} & 0.26\% & 0.13\% & 0.34\% & 0.16\% \\

\hline
\textbf{Data-driven} & 0.47\% & 0.34\% & 0.59\% & 0.39\%  \\
\hline
\end{tabular}
\label{table:DeepONet-resultsLow_Drag}
\end{table}

\begin{table}[H]
\caption{Mean and standard deviation of $L_1$ and $L_2$ relative errors between low-fidelity solution (data) and low-fidelity Fourier-featured deep operator networks' predictions for Lift coefficient, $C_L$.}
\centering

\begin{tabular}{c | c  c  c  c} 
\hline
\textbf{Low-fidelity model}  &\textbf{mean~  $L_1$}&\textbf{st.dev.~$L_1$}&\textbf{{mean~  $L_2$}}&\textbf{st.dev.~$L_2$} \\
\hline
\hline

\textbf{Physics-guided} & 0.71\% & 0.81\% & 0.90\% & 0.93\% \\
\hline
\textbf{Data-driven} & 0.88\% & 1.04\% & 1.06\% & 1.14\%  \\
\hline
\end{tabular}
\label{table:DeepONet-resultsLow_Lift}
\end{table}

\subsection{Bi-Fidelity Fourier-Featured Deep Operator Learning Results} \label{subsec:hig-fidelity-results}
In our proposed bi-fidelity learning framework, both the low-fidelity and residual models can be either physics-guided or data-driven  Fourier-featured DeepONet. To rigorously assess the effectiveness of this bi-fidelity structure and the performance of the introduced physics-guided Fourier-featured DeepONet, we constructed three distinct bi-fidelity configurations:

\begin{itemize}

\item \textbf{First Configuration:} Both the low-fidelity and the residual models are physics-guided Fourier-featured DeepONets.

\item \textbf{Second Configuration:}  The low-fidelity model is a data-driven  Fourier-featured DeepONet, while the residual model is a physics-guided Fourier-featured DeepONet.

\item \textbf{Third Configuration:} Both the low-fidelity and residual models are data-driven  Fourier-featured DeepONets.
\end{itemize}
Through these diverse configurations, we aim to provide a comprehensive analysis of the data-driven and physics-guided approaches within our bi-fidelity learning framework.

This section focuses on the assessment of these frameworks through their predictions on the high-fidelity test dataset. For this purpose, after training both the low-fidelity and residual DeepONets, we employ the methodology outlined in Algorithm~\ref{alg:bi-fidelity-learning} and Algorithm~\ref{alg:bi-fidelity-learningP} to approximate the high-fidelity solution for each framework. These solutions correspond with the respective bi-fidelity frameworks' predictions on the high-fidelity test dataset.

\begin{figure}[!h] 
    \begin{subfigure}[!bl]{0.45\textwidth}
        \begin{tikzpicture}[spy using outlines={magnification=2, width=3cm, height=0.8cm, connect spies}]
        \begin{axis}[width=\linewidth, xlabel={$t (s)$}, ylabel={$C_D$}, xmin=1, xmax=8, ymin=3.2, ymax=3.7, 
        xtick={0,1,...,10}, ytick={0,0.1,...,4}, legend pos=south east,]    
        \addplot [smooth, red] table{data/HFTrue_drag_PhysicsPhysics.txt}; 
        \label{ClTrue};    \addlegendentry[font=\scriptsize,]{High-Fidelity Data}
        \addplot [dashed, blue] table{data/HFPredic_drag_PhysicsPhysics.txt}; 
        \label{ClPredic}; \addlegendentry[font=\scriptsize,]{Bi-Fidelity Framework}
        \addplot [smooth, black] table{data/HFPredicLF_drag_PhysicsPhysics.txt}; 
        \label{ClPredic}; \addlegendentry[font=\scriptsize,]{Low-Fidelity DeepONet}
        \coordinate (spypoint) at (axis cs:5.3, 3.65);
        \coordinate (magnifyglass) at (axis cs:4.25, 3.76); 
        \end{axis}
        \spy on (spypoint) in node [fill=white] at (magnifyglass);
        \node at (0.5, 3.4) {\textbf{(a)}};
        \end{tikzpicture}
    \end{subfigure}
    \hspace{5pt}
    \begin{subfigure}[!br]{0.45\textwidth}
        \begin{tikzpicture}[spy using outlines={magnification=2, width=3cm, height=1cm, connect spies}]
        \begin{axis}[width=\linewidth, xlabel={$t (s)$},ylabel={$C_L$}, xmin=0, xmax=8, ymin=-1.3, ymax=1.3, 
        xtick={0,1,...,10}, ytick={-1.2,-0.6,...,1.2}, ] 
        \addplot [smooth, red] table{data/HFTrue_lift_PhysicsPhysics.txt}; 
        \label{ClTrue};             
        \addplot [dashed, blue] table{data/HFPredic_lift_PhysicsPhysics.txt}; 
        \label{ClPredic};           
        \addplot [smooth, black] table{data/HFPredicLF_lift_PhysicsPhysics.txt}; 
        \label{ClPredic};           
        \coordinate (spypoint) at (axis cs:4.5, 1.02);
        \coordinate (magnifyglass) at (axis cs:3.75, 1.7); 
        \end{axis}
        \spy on (spypoint) in node [fill=white] at (magnifyglass);
        \node at (0.5, 3.4) {\textbf{(b)}};
        \end{tikzpicture}
    \end{subfigure}
\caption{Comparison between the high-fidelity solution (data) time trajectory and the prediction from the first configuration of the bi-fidelity learning framework: (\textbf{Low-fidelity model: physics-guided Fourier-featured DeepONet; High-fidelity model: physics-guided Fourier-featured DeepONet}). Displayed for a random undisturbed free-stream velocity~($\overline{u}=1.06~m/s$): a) Drag Coefficient, $C_D$, and b) Lift Coefficient, $C_L$.}
\label{DragLiftHPP}
\end{figure}
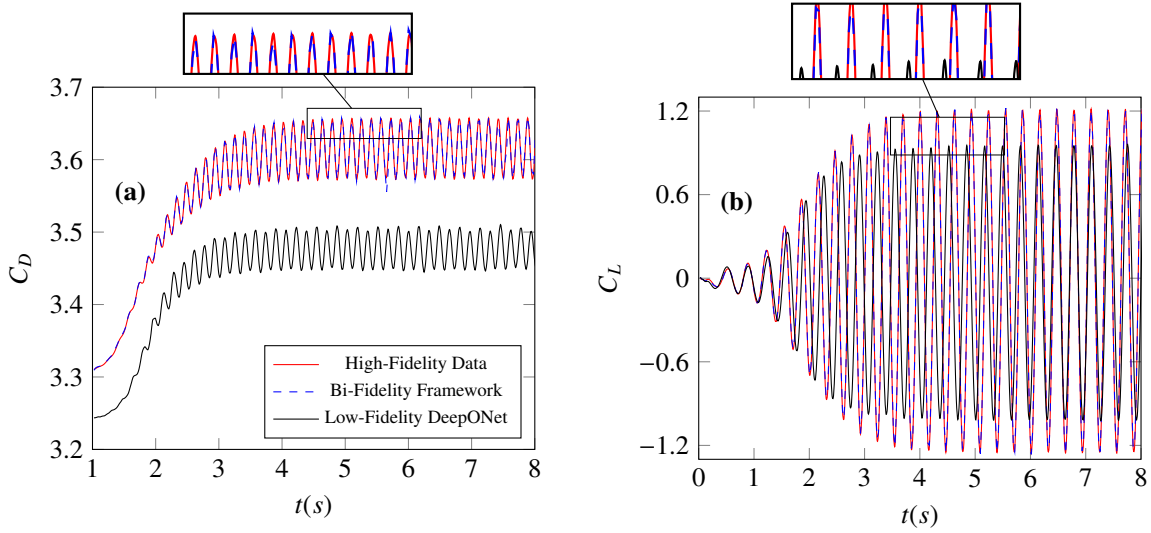

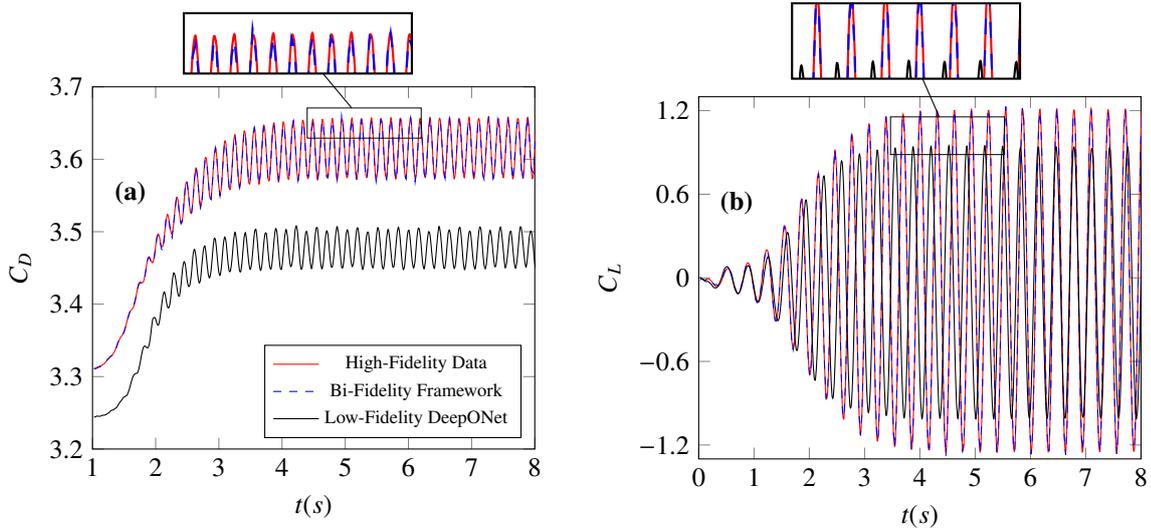
\begin{figure}[!h] 
    \begin{subfigure}[!bl]{0.45\textwidth}
        \begin{tikzpicture}[spy using outlines={magnification=2, width=3cm, height=0.8cm, connect spies}]
        \begin{axis}[width=\linewidth, xlabel={$t (s)$},ylabel={$C_D$}, xmin=1, xmax=8, ymin=3.2, ymax=3.7, 
        xtick={0,1,...,10}, ytick={0,0.1,...,4}, legend pos=south east,]    
        \addplot [smooth, red] table{data/HFTrue_drag_PhysicsData.txt}; 
        \label{ClTrue}; \addlegendentry[font=\scriptsize,]{High-Fidelity Data}
        \addplot [dashed, blue] table{data/HFPredic_drag_PhysicsData.txt}; 
        \label{ClPredic}; \addlegendentry[font=\scriptsize,]{Bi-Fidelity Framework}
        \addplot [smooth, black] table{data/HFPredicLF_drag_PhysicsData.txt}; 
        \label{ClPredic}; \addlegendentry[font=\scriptsize,]{Low-Fidelity DeepONet}
        \coordinate (spypoint) at (axis cs:5.3, 3.65);
        \coordinate (magnifyglass) at (axis cs:4.25, 3.76); 
        \end{axis}
        \spy on (spypoint) in node [fill=white] at (magnifyglass);
        \node at (0.5, 3.4) {\textbf{(a)}};
        \end{tikzpicture}
    \end{subfigure}
    \hspace{5pt}
    \begin{subfigure}[!br]{0.45\textwidth}
        \begin{tikzpicture}[spy using outlines={magnification=2, width=3cm, height=1cm, connect spies}]
        \begin{axis}[width=\linewidth, xlabel={$t (s)$},ylabel={$C_L$}, xmin=0, xmax=8, ymin=-1.3, ymax=1.3, 
        xtick={0,1,...,10}, ytick={-1.2,-0.6,...,1.2}, ] 
        \addplot [smooth, red] table{data/HFTrue_lift_PhysicsData.txt}; 
        \label{ClTrue};             
        \addplot [dashed, blue] table{data/HFPredic_lift_PhysicsData.txt}; 
        \label{ClPredic};           
        \addplot [smooth, black] table{data/HFPredicLF_lift_PhysicsData.txt}; 
        \label{ClPredic};           
        \coordinate (spypoint) at (axis cs:4.5, 1.02);
        \coordinate (magnifyglass) at (axis cs:3.75, 1.7); 
        \end{axis}
        \spy on (spypoint) in node [fill=white] at (magnifyglass);
        \node at (0.5, 3.4) {\textbf{(b)}};
        \end{tikzpicture}
    \end{subfigure}
\caption{Comparison between the high-fidelity solution (data) time trajectory and the prediction from the second configuration of the bi-fidelity learning framework: (\textbf{Low-fidelity model: data-driven  Fourier-featured DeepONet; High-fidelity model: physics-guided Fourier-featured DeepONet}). Displayed for a random undisturbed free-stream velocity~($\overline{u}=1.06~m/s$): a) Drag Coefficient, $C_D$, and b) Lift Coefficient, $C_L$.}
\label{DragLiftHDP}
\end{figure}

\begin{figure}[!h] 
    \begin{subfigure}[!bl]{0.45\textwidth}
        \begin{tikzpicture}[spy using outlines={magnification=2, width=3cm, height=0.8cm, connect spies}]
        \begin{axis}[width=\linewidth, xlabel={$t (s)$},ylabel={$C_D$}, xmin=1, xmax=8, ymin=3.2, ymax=3.7, 
        xtick={0,1,...,10}, ytick={0,0.1,...,4}, legend pos=south east,]    
        \addplot [smooth, red] table{data/HFTrue_drag_DataData.txt}; 
        \label{ClTrue};  \addlegendentry[font=\scriptsize,]{High-Fidelity Data}
        \addplot [dashed, blue] table{data/HFPredic_drag_DataData.txt}; 
        \label{ClPredic};  \addlegendentry[font=\scriptsize,]{Bi-Fidelity Framework}
        \addplot [smooth, black] table{data/HFPredicLF_drag_DataData.txt}; 
        \label{ClPredic}; \addlegendentry[font=\scriptsize,]{Low-Fidelity DeepONet}
        \coordinate (spypoint) at (axis cs:5.3, 3.65);
        \coordinate (magnifyglass) at (axis cs:4.25, 3.76); 
        \end{axis}
        \spy on (spypoint) in node [fill=white] at (magnifyglass);
        \node at (0.5, 3.4) {\textbf{(a)}};
        \end{tikzpicture}
    \end{subfigure}
    \hspace{5pt}
    \begin{subfigure}[!br]{0.45\textwidth}
        \begin{tikzpicture}[spy using outlines={magnification=2, width=3cm, height=1cm, connect spies}]
        \begin{axis}[width=\linewidth, xlabel={$t (s)$},ylabel={$C_L$}, xmin=0, xmax=8, ymin=-1.3, ymax=1.3, 
        xtick={0,1,...,10}, ytick={-1.2,-0.6,...,1.2}, ] 
        \addplot [smooth, red] table{data/HFTrue_lift_DataData.txt}; 
        \label{ClTrue};             
        \addplot [dashed, blue] table{data/HFPredic_lift_DataData.txt}; 
        \label{ClPredic};           
        \addplot [smooth, black] table{data/HFPredicLF_lift_DataData.txt}; 
        \label{ClPredic};           
        \coordinate (spypoint) at (axis cs:4.5, 1.02);
        \coordinate (magnifyglass) at (axis cs:3.75, 1.7); 
        \end{axis}
        \spy on (spypoint) in node [fill=white] at (magnifyglass);
        \node at (0.5, 3.4) {\textbf{(b)}};
        \end{tikzpicture}
    \end{subfigure}
\caption{Comparison between the high-fidelity solution (data) time trajectory and the prediction from the third configuration of the bi-fidelity learning framework: (\textbf{Low-fidelity model: data-driven  Fourier-featured DeepONet; High-fidelity model: data-driven  Fourier-featured DeepONet}). Displayed for a random undisturbed free-stream velocity~($\overline{u}=1.06~m/s$): a) Drag Coefficient, $C_D$, and b) Lift Coefficient, $C_L$.}
\label{DragLiftHDD}
\end{figure}
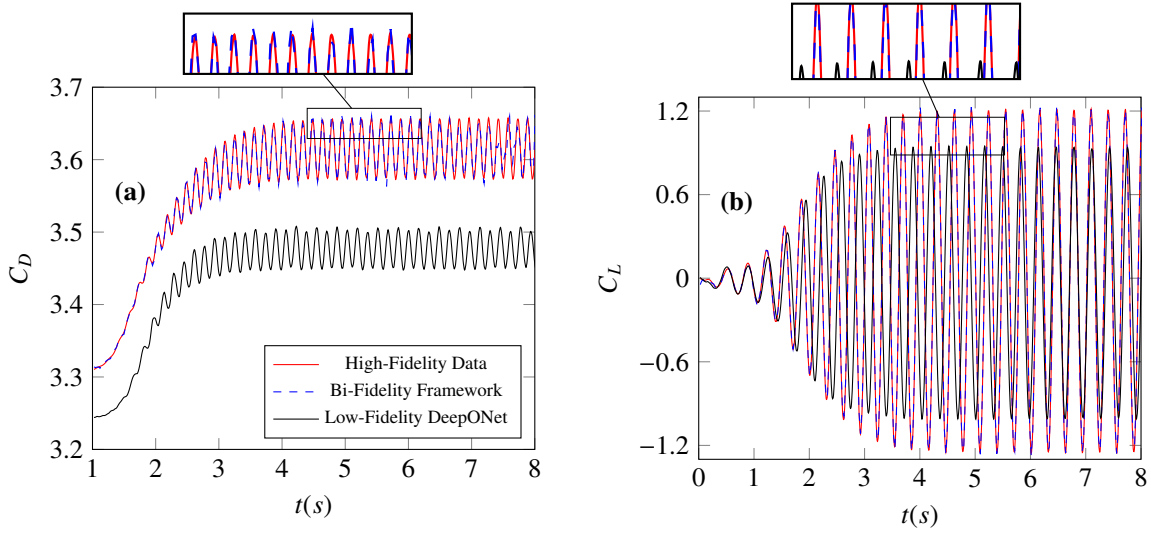

To demonstrate the performance of our proposed bi-fidelity learning framework, we randomly selected an undisturbed free-stream velocity \( u_j \in \mathcal{U} \) from the high-fidelity test dataset. Utilizing this velocity, we predicted the time trajectory of both drag and lift coefficients by employing each of the three previously outlined bi-fidelity learning framework configurations.
Figure~\ref{DragLiftHPP}, Figure~\ref{DragLiftHDP}, and Figure~\ref{DragLiftHDD} present the drag and lift coefficients' time trajectory for the chosen velocity. Correspondingly, these figures represent predictions obtained from the first, second, and third configurations, respectively. As observed in these figures, the fine solution for both drag and lift coefficients aligns closely with that of the bi-fidelity Fourier-featured deep operator learning frameworks' predictions, indicative of a strong match. Such observations imply not only the superior capability of the residual deep operator network in discerning the residuals between coarse predictions and fine solutions but also underscore the outstanding performance of our proposed bi-fidelity learning framework.

\subsubsection{Comparative Performance Analysis of Bi-Fidelity Learning Frameworks} 
To comprehensively evaluate the generalization performance of the proposed bi-fidelity learning frameworks, Tables~\ref{table:DeepONet-resultsLow_DragBi} and ~\ref{table:DeepONet-resultsLow_LiftBi} show the \(L_1\) and \(L_2\)-relative errors corresponding to each framework's predictions of the fine solution. These statistics underscore that all three bi-fidelity configurations achieve errors below $3\%$ when predicting the fine solution for the target coefficients. Interestingly, the results indicate a superior predictive capability from the physics-guided approach, attributable to its leveraging of the system's physical equations. Among the configurations, the first, which employs physics-guided Fourier-featured DeepONet for both low-fidelity and residual models, yields the least error. Furthermore, the second configuration, utilizing the physics-guided Fourier-featured DeepONet for the residual model and data-driven DeepONet for the low-fidelity model, exhibits lower errors than the third configuration that relies on data-driven  Fourier-featured DeepONet for both models. This clearly showcases the efficacy of the proposed physics-guided deep operator for both low-fidelity and residual models in the bi-fidelity framework.

\begin{table}[h!]
\caption{Mean and standard deviation of $L_1$ and $L_2$ relative errors between high-fidelity solution (data) and the bi-fidelity Fourier-featured deep operators learning frameworks' predictions for the Drag coefficient, $C_D$.}
\centering

\begin{tabular}{c  c | c  c  c  c} 
\hline
\textbf{Low-fidelity model}  & \textbf{Residual model}  &\textbf{mean~  $L_1$}&\textbf{st.dev.~$L_1$}&\textbf{{mean~  $L_2$}}&\textbf{st.dev.~$L_2$} \\
\hline
\hline

\textbf{Physics-guided} & \textbf{Physics-guided} & 0.83\% & 0.49\% & 1.19\% & 0.65\% \\
\hline
\textbf{Data-driven} & \textbf{Physics-guided} & 0.84\% & 0.46\% & 1.22\% & 0.63\%  \\
\hline
\textbf{Data-driven} & \textbf{Data-driven} & 1.24\% & 0.65\% & 1.73\% & 0.81\%  \\
\hline
\end{tabular}
\label{table:DeepONet-resultsLow_DragBi}
\end{table}

\begin{table}[h!]
\caption{Mean and standard deviation of $L_1$ and $L_2$ relative errors between high-fidelity solution (data) and the bi-fidelity Fourier-featured deep operators learning frameworks' predictions for Lift coefficient, $C_L$.}
\centering

\begin{tabular}{c  c | c  c  c  c} 
\hline
\textbf{Low-fidelity model}  & \textbf{Residual model}  &\textbf{mean~  $L_1$}&\textbf{st.dev.~$L_1$}&\textbf{{mean~  $L_2$}}&\textbf{st.dev.~$L_2$} \\
\hline
\hline

\textbf{Physics-guided} & \textbf{Physics-guided} & 1.60\% & 0.88\% & 2.02\% & 1.10\% \\
\hline
\textbf{Data-driven} & \textbf{Physics-guided} & 1.70\% & 0.75\% & 2.10\% & 0.92\%  \\
\hline
\textbf{Data-driven} & \textbf{Data-driven} & 2.03\% & 0.98\% & 2.55\% & 1.35\%  \\
\hline
\end{tabular}
\label{table:DeepONet-resultsLow_LiftBi}
\end{table}

\subsubsection{Comparison of Physics-Guided Bi-Fidelity Deep Operator Learning Framework: With vs. Without Fourier Feature  Trunk Network}
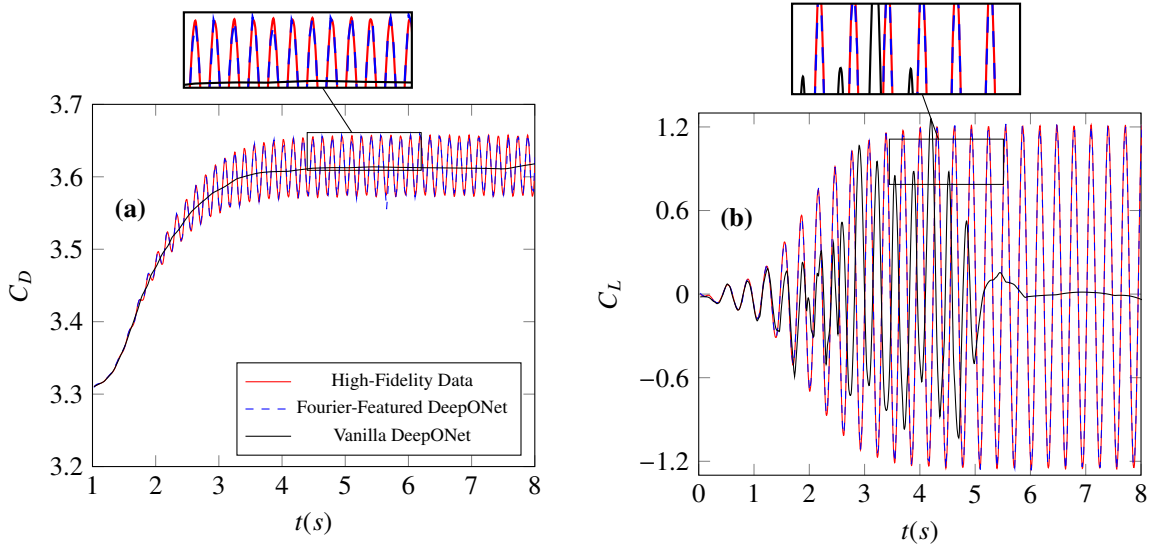
\begin{figure}[!h] 
    \begin{subfigure}[!bl]{0.45\textwidth}
        \begin{tikzpicture}[spy using outlines={magnification=2, width=3cm, height=1.0cm, connect spies}]
        \begin{axis}[width=\linewidth, xlabel={$t (s)$}, ylabel={$C_D$}, xmin=1, xmax=8, ymin=3.2, ymax=3.7, 
        xtick={0,1,...,10}, ytick={0,0.1,...,4}, legend pos=south east,]    
        \addplot [smooth, red] table{data/HFTrue_drag_PhysicsPhysics.txt}; 
        \label{ClTrue}; \addlegendentry[font=\scriptsize,]{High-Fidelity Data}
        \addplot [dashed, blue] table{data/HFPredic_drag_PhysicsPhysics.txt}; 
        \label{ClPredic}; \addlegendentry[font=\scriptsize,]{Fourier-Featured DeepONet}
        \addplot [smooth, black] table{data/HFPredic_drag_PhysicsPhysicslulu.txt}; 
        \label{ClPredic}; \addlegendentry[font=\scriptsize,]{Vanilla DeepONet}
        \coordinate (spypoint) at (axis cs:5.3, 3.635);
        \coordinate (magnifyglass) at (axis cs:4.25, 3.775); 
        \end{axis}
        \spy on (spypoint) in node [fill=white] at (magnifyglass);
        \node at (0.5, 3.4) {\textbf{(a)}};
        \end{tikzpicture}
    \end{subfigure}
    \hspace{5pt}
    \begin{subfigure}[!br]{0.45\textwidth}
        \begin{tikzpicture}[spy using outlines={magnification=2, width=3cm, height=1.2cm, connect spies}]
        \begin{axis}[width=\linewidth, xlabel={$t (s)$},ylabel={$C_L$}, xmin=0, xmax=8, ymin=-1.3, ymax=1.3, 
        xtick={0,1,...,10}, ytick={-1.2,-0.6,...,1.2}, ] 
        \addplot [smooth, red] table{data/HFTrue_lift_PhysicsPhysics.txt}; 
        \label{ClTrue};             
        \addplot [dashed, blue] table{data/HFPredic_lift_PhysicsPhysics.txt}; 
        \label{ClPredic};           
        \addplot [smooth, black] table{data/HFPredic_lift_PhysicsPhysicslulu.txt}; 
        \label{ClPredic};           
        \coordinate (spypoint) at (axis cs:4.48, 0.95);
        \coordinate (magnifyglass) at (axis cs:3.75, 1.76); 
        \end{axis}
        \spy on (spypoint) in node [fill=white] at (magnifyglass);
        \node at (0.5, 3.4) {\textbf{(b)}};
        \end{tikzpicture}
    \end{subfigure}
\caption{Comparison of physics-guided bi-fidelity deep operator learning frameworks utilizing: Vanilla DeepONet vs. Fourier-featured DeepONet. Displayed for a random undisturbed free-stream velocity~($\overline{u}=1.06~m/s$): a) Drag Coefficient, $C_D$, and b) Lift Coefficient, $C_L$.}
\label{DragLiftHPPlulu}
\end{figure}

As detailed in Section (\ref{subsubsec:Fourier-features-network}), we employed Fourier feature networks as the Trunk net for both the data-driven and physics-guided DeepONets in our low-fidelity and residual models. Referring to Figure \ref{DragLiftHPPlulu}, it becomes evident that the bi-fidelity deep operator learning framework constructed using the vanilla DeepONet struggles to capture the oscillatory characteristics of the lift and drag time trajectories. In contrast, the proposed bi-fidelity Fourier-featured deep operator learning framework adeptly captures the fluctuations of these trajectories. This underlines the effectiveness and appropriateness of integrating the Fourier-featured Trunk net with DeepONet, especially for lift and drag time evolution prediction.


\section{Discussion} \label{sec:discussion}
The developed physics-guided bi-fidelity Fourier-featured deep operator learning framework has demonstrated outstanding performance, achieving error rates below $2\%$ for both coarse and fine operator approximation. This indicates a highly accurate prediction of time trajectories for the drag and lift coefficients. This achievement is particularly noteworthy considering that only 45 high-fidelity input trajectories ($N=45$) were used in the training process. It highlights the high efficiency of the proposed framework, which effectively utilizes available information, such as inferred physics, especially in scenarios where obtaining high-fidelity data is expensive. These results clearly emphasize the potential of our framework in tackling complex computational challenges while optimizing resource utilization.

Based on our comprehensive comparative analysis, it is evident that the proposed physics-guided Fourier-featured DeepONet outperforms its data-driven counterpart, primarily due to its enhanced predictive power and efficiency in narrowing the solution space. This makes it an optimal tool within the bi-fidelity learning framework, particularly for the low-fidelity and residual models when predicting lift and drag coefficients. However, it's crucial to acknowledge several key distinctions between these two methodologies.

Physics-guided models are notable for their ability to incorporate underlying physics-based insights into their framework. This incorporation has the potential to generate more robust and accurate predictions. This feature is particularly advantageous when dealing with complex phenomena, such as predicting drag and lift coefficients across a wide range of input values. By integrating physics-based insights into the model, we can potentially improve the precision and reliability of predictions, even when dealing with a diverse and complex parameter space. 

On the other hand, while data-driven models are undoubtedly powerful, they may overlook intricate dependencies if these subtleties are not adequately captured within the dataset. Physics-guided models address this potential limitation by leveraging known relationships and behaviors, thereby reducing the risk of overlooking such details.

\subsection{Future Work} \label{future-work}
While recognizing the unique strengths and limitations of both data-driven and physics-guided approaches, our research establishes a foundation for future advancements in the bi-fidelity framework of lift and drag coefficients. This paves the way for exploring other bi-fidelity learning methods, such as input augmentation  \cite{lu2022multifidelity}. This approach has the potential to improve the accuracy of predicting fine solutions, offering a promising avenue for more precise and reliable predictions in bi-fidelity learning.

Furthermore, an exciting future direction involves expanding this paper from bi-fidelity to multi-fidelity and Bayesian operator learning~\cite{lin2023b,moya2023deeponet}. This expansion aims to optimize the use of data while mitigating the computational costs associated with predicting computationally expensive data. The ultimate goal is to enhance the model's prediction by maximizing generalization and accuracy. As a result, our research not only contributes to the current understanding but also lays the foundation for future innovations in predicting lift and drag coefficients' time trajectories.

Finally, we plan to use novel Bayesian multi-fidelity operator learning frameworks to optimize complex dynamical systems~\cite{SAHIN2024124813} or predict network dynamical systems~\cite{sun2023deepgraphonet} in a distributed and federated~\cite{2022Lin1} manner.

\section{Conclusion} \label{sec:Conclusion}
This paper presents a framework based on physics-guided bi-fidelity Fourier-featured deep operator learning. It aims to predict the time trajectories of lift and drag coefficients for a cylinder inside a channel. The proposed approach uses Deep Operator Neural Networks (DeepONet) to design a physics-guided strategy that approximates the operator mapping between undisturbed free-stream velocity and the corresponding drag and lift coefficients over time. The framework approximates both coarse and fine operators which map undisturbed free-stream velocity to the target time trajectories. The latter is achieved by integrating physics-guided low-fidelity and residual deep operator networks. To accurately capture the oscillatory nature inherent in these coefficients, our physics-guided strategy employs sinusoidal family functions, converting the prediction into an operator/functional inverse problem. Simultaneously, we incorporate the Fourier features network as the DeepONet's trunk, further enhancing the capture of these fluctuations through random Fourier mapping. To assess the predictive capabilities of the physics-guided approach against a data-driven strategy for lift and drag coefficients, we evaluated both methodologies using test datasets. We established three different configurations within our bi-fidelity framework and examined their performance on these test datasets. The results underscore that the physics-guided Fourier-featured DeepONet possesses an enhanced capacity for predicting the time trajectories of lift and drag, outperforming its data-driven counterparts. This proficiency directly enhances the accuracy of the bi-fidelity framework. Both low-fidelity and the proposed bi-fidelity framework effectively predict the drag and lift coefficients' time trajectories. As a result, our physics-guided bi-fidelity Fourier-featured framework offers improved performance with less high-fidelity data.

\section*{Acknowledgments}
IS gratefully acknowledges the support of Lillian Gilbreth Postdoctoral Fellowships from Purdue University's College of Engineering.

IR gratefully acknowledges the support of International Research Support Initiative Program (IRSIP), Higher Education Commission (HEC) of Pakistan.

GL gratefully acknowledges the support of the National Science Foundation (DMS-2053746, DMS-2134209, ECCS-2328241, and OAC-2311848), and U.S. Department of Energy (DOE) Office of Science Advanced Scientific Computing Research program DE-SC0021142 and DE-SC0023161.

\hspace{10in}
\hfill 

\newpage

\newpage

\bibliographystyle{elsarticle-num} 
\bibliography{refpaper.bib}

\begin{thebibliography}{10}
\expandafter\ifx\csname url\endcsname\relax
  \def\url#1{\texttt{#1}}\fi
\expandafter\ifx\csname urlprefix\endcsname\relax\def\urlprefix{URL }\fi
\expandafter\ifx\csname href\endcsname\relax
  \def\href#1#2{#2} \def\path#1{#1}\fi

\bibitem{Goodfellow-et-al-2016}
I.~Goodfellow, Y.~Bengio, A.~Courville, Deep Learning, MIT Press, 2016,
  \url{http://www.deeplearningbook.org}.

\bibitem{HAN19891619}
J.~Han, S.~Ou, J.~Park, C.~Lei, Augmented heat transfer in rectangular channels
  of narrow aspect ratios with rib turbulators, International Journal of Heat
  and Mass Transfer 32~(9) (1989) 1619--1630.
\newblock \href {https://doi.org/https://doi.org/10.1016/0017-9310(89)90044-6}
  {\path{doi:https://doi.org/10.1016/0017-9310(89)90044-6}}.

\bibitem{Paniagua2004578}
G.~Paniagua, R.~De´nos, S.~Almeida, {Effect of the Hub Endwall Cavity Flow on
  the Flow-Field of a Transonic High-Pressure Turbine}, Journal of
  Turbomachinery 126~(4) (2004) 578--586.
\newblock \href {https://doi.org/10.1115/1.1791644}
  {\path{doi:10.1115/1.1791644}}.

\bibitem{Sahin202007}
I.~Sahin, A.~Chen, C.-C. Shiau, J.-C. Han, R.~Krewinkel, Effect of 45-deg rib
  orientations on heat transfer in a rotating two-pass channel with aspect
  ratio from 4:1 to 2:1, ASME Journal of Turbomachinery 142~(7) (2020) 071003,
  https://doi.org/10.1115/1.4046492.

\bibitem{Sahin202107}
I.~Sahin, I.-L. Chen, L.~M. Wright, J.-C. Han, H.~Xu, M.~Fox, Heat transfer in
  rotating, trailing edge, converging channels with smooth and pin-fins, ASME
  Journal of Turbomachinery 143~(7) (2021) 071007,
  https://doi.org/10.1115/1.4050355.

\bibitem{LEIFSSON201098}
L.~Leifsson, S.~Koziel, Multi-fidelity design optimization of transonic
  airfoils using physics-based surrogate modeling and shape-preserving response
  prediction, Journal of Computational Science 1~(2) (2010) 98--106.
\newblock \href {https://doi.org/https://doi.org/10.1016/j.jocs.2010.03.007}
  {\path{doi:https://doi.org/10.1016/j.jocs.2010.03.007}}.

\bibitem{LEIFSSON201545}
L.~Leifsson, S.~Koziel,
  \href{https://www.sciencedirect.com/science/article/pii/S1877750315000071}{Aerodynamic
  shape optimization by variable-fidelity computational fluid dynamics models:
  A review of recent progress}, Journal of Computational Science 10 (2015)
  45--54.
\newblock \href {https://doi.org/https://doi.org/10.1016/j.jocs.2015.01.003}
  {\path{doi:https://doi.org/10.1016/j.jocs.2015.01.003}}.
\newline\urlprefix\url{https://www.sciencedirect.com/science/article/pii/S1877750315000071}

\bibitem{SEN2018434}
O.~Sen, N.~J. Gaul, K.~Choi, G.~Jacobs, H.~Udaykumar,
  \href{https://www.sciencedirect.com/science/article/pii/S0021999118303486}{Evaluation
  of multifidelity surrogate modeling techniques to construct closure laws for
  drag in shock–particle interactions}, Journal of Computational Physics 371
  (2018) 434--451.
\newblock \href {https://doi.org/https://doi.org/10.1016/j.jcp.2018.05.039}
  {\path{doi:https://doi.org/10.1016/j.jcp.2018.05.039}}.
\newline\urlprefix\url{https://www.sciencedirect.com/science/article/pii/S0021999118303486}

\bibitem{DU2019371}
X.~Du, J.~Ren, L.~Leifsson, Aerodynamic inverse design using multifidelity
  models and manifold mapping, Aerospace Science and Technology 85 (2019)
  371--385.
\newblock \href {https://doi.org/https://doi.org/10.1016/j.ast.2018.12.008}
  {\path{doi:https://doi.org/10.1016/j.ast.2018.12.008}}.

\bibitem{Ertan2022}
E.~Demiral, C.~Sahin, K.~Arslan,
  \href{https://arc.aiaa.org/doi/abs/10.2514/6.2022-4161}{Aerodynamic Shape
  Optimization Using Multi-fidelity Surrogate-based Approach for
  Computationally Expensive Problems}.
\newblock \href
  {http://arxiv.org/abs/https://arc.aiaa.org/doi/pdf/10.2514/6.2022-4161}
  {\path{arXiv:https://arc.aiaa.org/doi/pdf/10.2514/6.2022-4161}}, \href
  {https://doi.org/10.2514/6.2022-4161} {\path{doi:10.2514/6.2022-4161}}.
\newline\urlprefix\url{https://arc.aiaa.org/doi/abs/10.2514/6.2022-4161}

\bibitem{Robinson20067114}
T.~Robinson, K.~Willcox, M.~Eldred, R.~Haimes, Multifidelity Optimization for
  Variable-Complexity Design.
\newblock \href {https://doi.org/10.2514/6.2006-7114}
  {\path{doi:10.2514/6.2006-7114}}.

\bibitem{HUANG2013279}
L.~Huang, Z.~Gao, D.~Zhang,
  \href{https://www.sciencedirect.com/science/article/pii/S1000936113000162}{Research
  on multi-fidelity aerodynamic optimization methods}, Chinese Journal of
  Aeronautics 26~(2) (2013) 279--286.
\newblock \href {https://doi.org/https://doi.org/10.1016/j.cja.2013.02.004}
  {\path{doi:https://doi.org/10.1016/j.cja.2013.02.004}}.
\newline\urlprefix\url{https://www.sciencedirect.com/science/article/pii/S1000936113000162}

\bibitem{Daniel2008}
D.~Jaeggi, G.~Parks, W.~Dawes, J.~Clarkson,
  \href{https://arc.aiaa.org/doi/abs/10.2514/6.2008-6052}{Robust Multi-Fidelity
  Aerodynamic Design Optimization Using Surrogate Models}.
\newblock \href
  {http://arxiv.org/abs/https://arc.aiaa.org/doi/pdf/10.2514/6.2008-6052}
  {\path{arXiv:https://arc.aiaa.org/doi/pdf/10.2514/6.2008-6052}}, \href
  {https://doi.org/10.2514/6.2008-6052} {\path{doi:10.2514/6.2008-6052}}.
\newline\urlprefix\url{https://arc.aiaa.org/doi/abs/10.2514/6.2008-6052}

\bibitem{HAN2013177}
Z.-H. Han, S.~Görtz, R.~Zimmermann, Improving variable-fidelity surrogate
  modeling via gradient-enhanced kriging and a generalized hybrid bridge
  function, Aerospace Science and Technology 25~(1) (2013) 177--189.
\newblock \href {https://doi.org/https://doi.org/10.1016/j.ast.2012.01.006}
  {\path{doi:https://doi.org/10.1016/j.ast.2012.01.006}}.

\bibitem{Vapnik201516}
V.~N. Vapnik, R.~Izmailov, Learning using privileged information: similarity
  control and knowledge transfer, J. Mach. Learn. Res. 16 (2015) 2023--2049.

\bibitem{Dehghani2017}
M.~Dehghani, A.~Mehrjou, S.~Gouws, J.~Kamps, B.~Sch{\"o}lkopf,
  Fidelity-weighted learning, ArXiv abs/1711.02799 (2017).

\bibitem{De2020543}
S.~De, J.~Britton, M.~Reynolds, R.~Skinner, K.~Jansen, A.~Doostan, On transfer
  learning of neural networks using bi-fidelity data for uncertainty
  propagation, International Journal for Uncertainty Quantification 10~(6)
  (2020) 543--573.

\bibitem{Lu20223210}
L.~Lu, R.~Pestourie, S.~G. Johnson, G.~Romano, Multifidelity deep neural
  operators for efficient learning of partial differential equations with
  application to fast inverse design of nanoscale heat transport, Physical
  Review Research 4~(2) (2022) 023210.

\bibitem{De20231432}
S.~De, M.~Reynolds, M.~Hassanaly, R.~N. King, A.~Doostan, Bi-fidelity modeling
  of uncertain and partially unknown systems using deeponets, Computational
  Mechanics 71~(4) (2023) 1251--1217, 10.1007/s00466-023-02272-4.

\bibitem{moya2023approximating}
C.~Moya, G.~Lin, T.~Zhao, M.~Yue, On approximating the dynamic response of
  synchronous generators via operator learning: A step towards building deep
  operator-based power grid simulators, arXiv preprint arXiv:2301.12538 (2023).

\bibitem{lin2023learning}
G.~Lin, C.~Moya, Z.~Zhang, Learning the dynamical response of nonlinear
  non-autonomous dynamical systems with deep operator neural networks,
  Engineering Applications of Artificial Intelligence 125 (2023) 106689.

\bibitem{zhang2023bayesian}
Z.~Zhang, C.~Moya, W.~T. Leung, G.~Lin, H.~Schaeffer, Bayesian deep operator
  learning for homogenized to fine-scale maps for multiscale pde, arXiv
  preprint arXiv:2308.14188 (2023).

\bibitem{Schafer1996}
M.~Sch{\"a}fer, S.~Turek, F.~Durst, E.~Krause, R.~Rannacher, Benchmark
  Computations of Laminar Flow Around a Cylinder, Vieweg and Teubner Verlag,
  Wiesbaden, 1996, pp. 547--566.
\newblock \href {https://doi.org/10.1007/978-3-322-89849-439}
  {\path{doi:10.1007/978-3-322-89849-439}}.

\bibitem{karniadakis2021physics}
G.~E. Karniadakis, I.~G. Kevrekidis, L.~Lu, P.~Perdikaris, S.~Wang, L.~Yang,
  Physics-informed machine learning, Nature Reviews Physics 3~(6) (2021)
  422--440.

\bibitem{2021Lulu1}
L.~Lu, P.~Jin, G.~Pang, Z.~Zhang, G.~E. Karniadakis, Learning nonlinear
  operators via deeponet based on the universal approximation theorem of
  operators, Nature Machine Intelligence 3~(3) (2021) 218--229,
  https://doi.org/10.1038/s42256-021-00302-5.

\bibitem{1995Chen1}
T.~Chen, H.~Chen, Universal approximation to nonlinear operators by neural
  networks with arbitrary activation functions and its application to dynamical
  systems, IEEE Transactions on Neural Networks 6~(4) (1995) 911--917.
\newblock \href {https://doi.org/https://doi.org/10.1109/72.392253}
  {\path{doi:https://doi.org/10.1109/72.392253}}.

\bibitem{tancik2020fourier}
M.~Tancik, P.~Srinivasan, B.~Mildenhall, S.~Fridovich-Keil, N.~Raghavan,
  U.~Singhal, R.~Ramamoorthi, J.~Barron, R.~Ng, Fourier features let networks
  learn high frequency functions in low dimensional domains, Advances in Neural
  Information Processing Systems 33 (2020) 7537--7547.

\bibitem{lu2022multifidelity}
L.~Lu, R.~Pestourie, S.~G. Johnson, G.~Romano, Multifidelity deep neural
  operators for efficient learning of partial differential equations with
  application to fast inverse design of nanoscale heat transport, Physical
  Review Research 4~(2) (2022) 023210.

\bibitem{lin2023b}
G.~Lin, C.~Moya, Z.~Zhang, B-deeponet: An enhanced bayesian deeponet for
  solving noisy parametric pdes using accelerated replica exchange sgld,
  Journal of Computational Physics 473 (2023) 111713.

\bibitem{moya2023deeponet}
C.~Moya, S.~Zhang, G.~Lin, M.~Yue, Deeponet-grid-uq: A trustworthy deep
  operator framework for predicting the power grid’s post-fault trajectories,
  Neurocomputing (2023).

\bibitem{SAHIN2024124813}
I.~Sahin, C.~Moya, A.~Mollaali, G.~Lin, G.~Paniagua,
  \href{https://www.sciencedirect.com/science/article/pii/S0017931023009584}{Deep
  operator learning-based surrogate models with uncertainty quantification for
  optimizing internal cooling c\ hannel rib profiles}, International Journal of
  Heat and Mass Transfer 219 (2024) 124813.
\newblock \href
  {https://doi.org/https://doi.org/10.1016/j.ijheatmasstransfer.2023.124813}
  {\path{doi:https://doi.org/10.1016/j.ijheatmasstransfer.2023.124813}}.
\newline\urlprefix\url{https://www.sciencedirect.com/science/article/pii/S0017931023009584}

\bibitem{sun2023deepgraphonet}
Y.~Sun, C.~Moya, G.~Lin, M.~Yue, Deepgraphonet: A deep graph operator network
  to learn and zero-shot transfer the dynamic response of networked systems,
  IEEE Systems Journal (2023).

\bibitem{2022Lin1}
C.~Moya, G.~Lin, Fed-deeponet: Stochastic gradient-based federated training of
  deep operator networks, Algorithms 15~(9), https://doi.org/10.3390/a15090325
  (2022).

\end{thebibliography}

\end{document}